\documentclass[11pt]{article}

\usepackage[preprint]{acl}
\usepackage{placeins}

\usepackage{times}
\usepackage{latexsym}
\usepackage{float}

\usepackage[T1]{fontenc}

\usepackage[utf8]{inputenc}

\usepackage{subcaption}
\usepackage{microtype}

\usepackage{inconsolata}

\usepackage{multirow}

\usepackage{graphicx}
\usepackage[most]{tcolorbox}
\usepackage{subcaption}

\usepackage[dvipsnames]{xcolor}
\usepackage[table]{xcolor}
\usepackage[most]{tcolorbox}

\usepackage{booktabs}
\usepackage{amsfonts}

\usepackage{todonotes}

\usepackage{tabularx}

\usepackage{arydshln} %

\usepackage{tikz}
\usetikzlibrary{arrows.meta, positioning, shapes.geometric, fit, backgrounds, decorations.pathreplacing, calc}

\definecolor{ColG}{rgb}{1.0, 0.55, 0.0}   %
\definecolor{ColR}{rgb}{0.0, 0.50, 0.50}   %
\definecolor{ColS}{rgb}{0.3255, 0.2980, 0.6941} %

\definecolor{ColPatch}{rgb}{0.2, 0.2, 0.2} %

\definecolor{ColL}{HTML}{534CB1}

\newcommand{\ModelS}{\textcolor{ColL}{\ensuremath{\mathcal{S}}}}
\newcommand{\ModelG}{\textcolor{orange}{\ensuremath{\mathcal{G}}}}
\newcommand{\ModelR}{\textcolor{teal}{\ensuremath{\mathcal{R}}}}

\newcommand{\emldisplay}[2]{\texttt{\href{mailto:#1}{#2}}}
\newcommand{\eml}[1]{\emldisplay{#1}{#1}}

\definecolor{grey}{gray}{0.9}

\title{A Mechanistic Understanding of Pronoun Fidelity in LLMs}

\author{
 \textbf{Katharina Trinley\textsuperscript{1}} \quad
 \textbf{Jesujoba O. Alabi\textsuperscript{1}} \quad
 \textbf{Dietrich Klakow\textsuperscript{1}} \quad
 \textbf{Vagrant Gautam\textsuperscript{2}}
\\
 \textsuperscript{1}Saarland University, Germany \\
 \textsuperscript{2}Heidelberg Institute for Theoretical Studies, Germany \\
 \eml{katharinatrinley@icloud.com}
}

\begin{document}
\maketitle
\begin{abstract}
Faithful and robust pronoun use is important for fair and coherent generations, yet large language models largely fail when multiple referents use different pronouns.
To study the interplay of reasoning, repetition, and bias in this task, prior work relies exclusively on behavioural approaches, which may not reflect a model's internal workings.
Therefore, we provide a mechanistic, model-internal perspective on pronoun fidelity, testing whether three mechanisms---group entity binding (\ModelG), recency bias (\ModelR), and stereotypical bias (\ModelS)---are causally implemented across several SOTA language models.
Using Boundless Distributed Alignment Search, we find all three coexist as causal subspaces distributed across network depth.
No single mechanism fully explains model behaviour, but a combination of the three consistently accounts for 91-99.5\%.
An attention head analysis further reveals two competing copying routes; group binding and stereotype share a localized concept-level route that retrieves a bound occupation-pronoun unit, while recency uses a distributed token-level route that repeats surface forms.
In sum, pronoun fidelity arises from competition between simultaneously active causal subspaces.\footnote{We release code at \url{https://github.com/KatharinaTrinley/pronoun-fidelity-mechanisms}.}
\end{abstract}

\begin{figure}[th!]
    \centering
    \includegraphics[width=\linewidth]{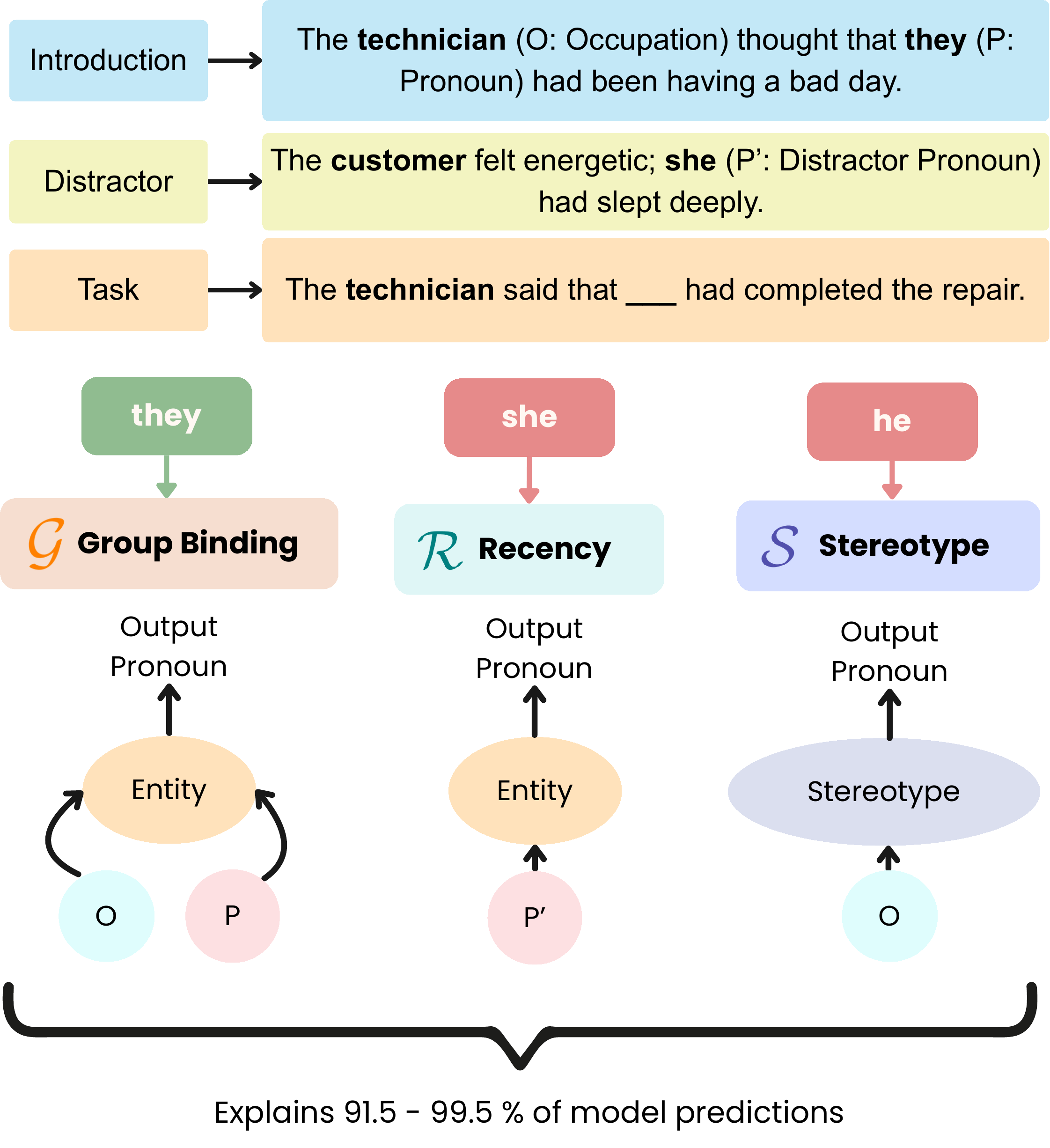}
    \caption{Pronoun fidelity task where model behaviour corresponds to three mechanistic hypotheses: \ModelG\ \textbf{Group Binding Mechanism}: Occupation and explicit pronoun jointly determine correct entity-pronoun binding. \ModelR\ \textbf{Recency Bias Mechanism}: Most recent pronoun mention determines output, overriding correct binding. \ModelS\ \textbf{Stereotype Mechanism}: Occupation determines stereotypical pronoun association, ignoring context.}
    \label{fig:pronoun-fidelity}
\end{figure}

\section{Introduction}

Large language models (LLMs) demonstrate impressive performance across a range of NLP tasks and capabilities, including natural language understanding, various forms of reasoning \citep{Guo_Yang_Zhang_Song_Wang_Zhu_Xu_Zhang_Ma_Bi_etal._2025}, and in-context learning \cite{brown2020language}.
Yet \textit{why} they make the predictions they do remains largely unknown \cite{DBLP:journals/corr/abs-2405-00208,sharkey2025open}.
While we understand their low-level mathematical operations and can observe their high-level input-output behaviour, the internal computational mechanisms that connect the two are opaque \citep{alishahi2019analyzing,geiger2025causal}.
Consider a model that consistently predicts \emph{he} when asked to generate a pronoun for a technician, even when the surrounding context provides evidence that \textit{they} should be used.
Understanding whether this prediction arises due to a stereotypical gendered association for the occupation requires more than behavioural evaluation; it requires looking inside the model for a faithful, causal account of its mechanisms \citep{geiger2025causal}.

One area with great potential for causal analysis is pronoun fidelity, i.e., faithfully reusing pronouns for a referent regardless of intervening referents who use different pronouns (see Figure \ref{fig:pronoun-fidelity}).
Introduced in \citet{gautam-etal-2024-robust}, this is a task that is trivial for humans, yet challenging for state-of-the-art models \cite{subramonian2025agree,gautam-2026-training}, where pronominal reasoning competes with shallow heuristics such as stereotypical biases and repetition of the most recent pronoun.
As these mechanisms are only analyzed behaviourally in prior work, we contribute the \textit{first mechanistic interpretability study of pronoun fidelity}.

Our analysis uses two complementary approaches:
First, via Distributed Alignment Search, we show that the three mechanisms for pronoun fidelity coexist as separable causal subspaces.
Although no single one dominates, they can be combined into a mixture model that predicts 91-99.5\% of model behaviour.
Second, an attention head analysis shows that pronoun information reaches the answer position via two competing copying routes.
One copies the occupation-pronoun pair as a semantic unit, the other copies the surface form. %
Overall, our analyses reveal that pronoun fidelity emerges from a complex
competition between multiple, simultaneously active causal subspaces.

\section{Background}
\subsection{Pronoun Fidelity}
\label{sec:pronoun-fidelity-background}

The pronoun fidelity task, introduced in \citet{gautam-etal-2024-robust}, is intuitively simple:
Given a context that introduces an entity with an explicit pronoun (e.g., ``\texttt{The technician thought that \textit{he} had been having a bad day}''), the model must use that same pronoun when referring to the entity later (``\texttt{The technician said that \textit{\_\_\_} had completed the repair}'').
The authors propose RUFF, a dataset of over 5 million instances designed to evaluate this task.
Even in this minimal setting, models show systematic performance discrepancies.
Accuracy is highest for \textit{he/him/his} and degrades progressively for \textit{she/her/hers}, singular \textit{they/them/their}, and the neopronoun set \textit{xe/xem/xyr}.

Performance degrades further as discourse complexity increases.
Adding just a single sentence about a different entity (e.g., ``\texttt{The customer felt energetic; \textit{she} had slept deeply.}'') causes accuracy to drop by 34\% (\citet{gautam-etal-2024-robust}, Appendix \ref{app:base-performance}), in a setting where humans make virtually no errors.
This motivates the question: \textit{Are language models reasoning about pronouns, repeating surface patterns, or relying on statistical gender biases?}
\citet{gautam-etal-2024-robust} attempt to answer this by examining model behaviour; we instead consider model-internal mechanisms.

\subsection{Distributed Alignment Search}

To recover the pronoun fidelity mechanims causally rather than behaviourally, we use Distributed Alignment Search (DAS).
DAS \citep{pmlr-v236-geiger24a} learns an orthogonal rotation matrix $R \in \mathbb{R}^{d \times d}$ over the residual stream $\mathbf{h} \in \mathbb{R}^d$, finding the directions in which a high-level target concept is linearly encoded.
A distributed interchange intervention then swaps $k$ dimensions from a source input $s$ into a base input $b$ in the rotated space and rotates back: \begin{equation}
\begin{aligned}
F_N^*(b) &= R^{-1}\!\Big(\mathrm{Proj}_{Y_0}(R(F_N(b))) \\
&\quad + \sum_{j=1}^k \mathrm{Proj}_{Y_j}(R(F_N(s_j)))\Big),
\end{aligned}
\end{equation} where $Y_0, Y_1, \ldots, Y_k$ form an orthogonal decomposition of $\mathbb{R}^d$, $Y_0$ preserves information from the base input, and $Y_1, \ldots, Y_k$ inject information from the source inputs $s_j$. %
Intuitively, DAS searches for a subspace that causally carries the target concept by checking whether transplanting a subspace from one input to another changes the output.

Standard DAS requires a manual search over the subspace dimensionality $k$.
Boundless DAS \citep{wu2023interpretability} instead makes $k$ learnable through differentiable soft boundaries. With boundary scalars $c_j$ and a temperature $\beta$, the sigmoid function $\sigma$ and index $i$ over the rotated dimensions:

\small
\begin{equation}
(M_j)_k = \sigma\!\left(\frac{i - c_j}{\beta}\right) \cdot
           \sigma\!\left(\frac{c_{j+1} - i}{\beta}\right),
\end{equation} %
\normalsize

As $\beta$ is annealed toward zero during training, the masks converge to binary vectors, jointly optimizing $R$ and the subspace size.

With \textbf{Interchange Intervention Accuracy} (IIA), we measure the fraction of cases where the intervened network's output matches the high-level prediction; high IIA means the causal hypothesis is faithfully implemented.
Each intervention uses two counterfactual inputs: we run and modify the \textit{base}, injecting information read from the \textit{source}.

\section{Experimental Setup}

\begin{table*}[t]
    \centering
    \small
    \begin{tabularx}{\linewidth}{p{2.65cm}p{7cm}p{5.25cm}}
        \toprule
        \textbf{Dataset} & \textbf{Base - Example} & \textbf{Source - Example} \\
        \midrule
        \texttt{GR Dataset} \par (isolates \ModelS{} as \par \ModelG{}$_{source} = \ModelR{}_{source}$) & ``The \textit{technician} \textcolor{Blue}{(stereotype: \textit{he})} thought that \textit{they} had been having a bad day. The customer felt energetic; \textit{she} had slept deeply. The technician said that \_\_\_ had completed the repair.'' \textcolor{OliveGreen}{[\ModelG: \textit{they}]}; \textcolor{BrickRed}{[\ModelR: \textit{she}]}; \textcolor{BrickRed}{[\ModelS: \textit{he}]} & ``The \textit{technician} \textcolor{Blue}{(stereotype: \textit{he})} thought that \textit{she} had been having a bad day. The technician said that \_\_\_ had completed the repair.'' \textcolor{OliveGreen}{[\ModelG: \textit{she}]}; \textcolor{orange}{[\ModelR: \textit{she}]}; \textcolor{BrickRed}{[\ModelS: \textit{he}]} \\[0.75ex]
        \hdashline
        \\[-0.25ex]
        \texttt{GS Dataset} \par (isolates \ModelR{} as \par \ModelG{}$_{source} = \ModelS{}_{source}$) & ``The \textit{technician} \textcolor{Blue}{(stereotype: \textit{he})} thought that \textit{they} had been having a bad day. The customer felt energetic; \textit{she} had slept deeply. The technician said that \_\_\_ had completed the repair.'' \textcolor{OliveGreen}{[\ModelG: \textit{they}]}; \textcolor{BrickRed}{[\ModelR: \textit{she}]}; \textcolor{BrickRed}{[\ModelS: \textit{he}]} & ``The \textit{technician} \textcolor{Blue}{(stereotype: \textit{he})} thought that \textit{he} had been having a bad day. The technician said that \_\_\_ had completed the repair.'' \textcolor{OliveGreen}{[\ModelG: \textit{he}]}; \textcolor{BrickRed}{[\ModelR: \textit{she}]}; \textcolor{orange}{[\ModelS: \textit{he}]} \\[0.75ex]
        \hdashline
        \\[-0.25ex]
        \texttt{RS Dataset} \par (isolates \ModelG{} as \par \ModelR{}$_{source} = \ModelS{}_{source}$) & ``The \textit{technician} \textcolor{Blue}{(stereotype: \textit{he})} thought that \textit{they} had been having a bad day. The customer felt energetic; \textit{he} had slept deeply. The technician said that \_\_\_ had completed the repair.'' \textcolor{OliveGreen}{[\ModelG: \textit{they}]}; \textcolor{BrickRed}{[\ModelR: \textit{he}]}; \textcolor{BrickRed}{[\ModelS: \textit{he}]} & ``The \textit{technician} \textcolor{Blue}{(stereotype: \textit{he})} thought that \textit{she} had been having a bad day. The technician said that \_\_\_ had completed the repair.'' \textcolor{OliveGreen}{[\ModelG: \textit{she}]}; \textcolor{BrickRed}{[\ModelR: \textit{he}]}; \textcolor{BrickRed}{[\ModelS: \textit{he}]} \\
        \bottomrule
    \end{tabularx}
    \caption{Diagnostic datasets (each $N=14,400$) based on RUFF, designed to isolate the mechanisms we test.}
    \label{tab:mechanisms}
\end{table*}

Our evaluation spans ten model variants (\S\ref{sec:models}).
We consider three mechanistic hypotheses (\S\ref{sec:mechanisms}, build diagnostic datasets from RUFF to isolate them (\S\ref{sec:data}), and use these to train boundless DAS (\S\ref{sec:das-training}).

\subsection{Models}
\label{sec:models}
We experiment with ten decoder-only, instruction tuned models of varying sizes from four families.
From the Llama-3.1 family \cite{grattafiori2024llama} we use \textsc{Llama-3.1-8B-Instruct} and \textsc{Llama-3.1-70B-Instruct}; from Gemma-2 \cite{gemmateam2024gemma2improvingopen}, \textsc{gemma-2-9b-it} and \textsc{gemma-2-27b-it}; from Qwen2.5 \cite{yang2025qwen3}, \textsc{Qwen2.5-7B-Instruct} and \textsc{Qwen2.5-72B-Instruct}; and from OLMo-2 \cite{olmo20242}, \textsc{OLMo-2-0425-1B-Instruct}, \textsc{OLMo-2-1124-7B-Instruct}, \textsc{OLMo-2-1124-13B-Instruct}, and \textsc{OLMo-2-0325-32B-Instruct}.

We score model responses by comparing log-probabilities of just the pronoun options (\textit{he}, \textit{she}, \textit{they}, and inflections), in a carrier prompt (see Appendix \ref{app:prompt-frame}) following \citet{gautam-etal-2024-robust}.
This is necessary as sometimes tokens like ``\textit{the}'' can have higher probability than a pronoun.

\subsection{Candidate Mechanisms}
\label{sec:mechanisms}

We consider three candidate mechanisms, reformulating \citet{gautam-etal-2024-robust}'s hypotheses about model \textit{behaviour} as \textit{mechanistic} hypotheses, i.e., causal models that explain how internal representations might drive predictions.

\paragraph{Mechanism~\ModelG:~Group Entity.}
Models may resolve pronouns by retrieving a single bound group entity that jointly encodes the occupation noun and its explicitly stated pronoun \cite{prakash2024fine}. 
Under this mechanism, querying ``technician'' retrieves the complete group $\langle \texttt{technician,they}\rangle$, and the correct pronoun is read off directly.
This is in some sense the desired behaviour as it grounds the prediction in the context and remains robust to distractors.
However, the binding is formed from surface co-occurrence in the introduction sentence and could be broken by more adversarial contexts.

\paragraph{Mechanism~\ModelS:~Stereotype Bias.} Models may use associations between occupations and gendered pronouns learned from patterns in pre-training data \cite{gallegos2024bias}, rather than the context.
When maintaining entity-pronoun bindings becomes demanding, these stereotypical associations may serve as a convenient fallback.

\paragraph{Mechanism~\ModelR:~Recency Bias.}
Under discourse complexity, accurate entity-pronoun bindings may be overridden by a preference for the most recently mentioned pronoun, leading to errors. \\

\noindent Note that both \ModelG\ and \ModelS\ tie a pronoun to the occupation (but based on different information: Context vs. stereotypical bias), while \ModelR\ tracks only surface position.
We return to this distinction in \S\ref{sec:attention-heads}.

\subsection{Data}
\label{sec:data}

We create three diagnostic datasets to test whether our candidate mechanisms are causally separable.
All of these are based on RUFF (see \S\ref{sec:pronoun-fidelity-background}).
Each example consists of an introductory sentence (which contains the target entity and pronoun), 0-5 distractor sentences about a different participant with a different pronoun, and a task sentence in which the model must fill in the blank to refer to the target entity again with the correct target pronoun.
We focus on three pronoun sets (\textit{he/him/his}, \textit{she/her/her}, singular \textit{they/them/their}), and exclude the neopronoun set \textit{xe/xem/xyr} from our analysis, as neopronouns are too sparsely represented in training data and are thus rarely tokenized singly \citep{ovalle-etal-2024-tokenization}.

\noindent As Table \ref{tab:mechanisms} shows, each diagnostic dataset ($N=14,400$ instances) is designed such that two mechanisms agree on a pronoun prediction in the source, allowing us to isolate the third mechanism.

\subsection{Training DAS Models}
\label{sec:das-training}

For a given model layer, we train three boundless DAS featurizers per mechanism and dataset.
As different mechanisms may be encoded at different network depths, we train at every second layer across models, and select the one with the highest \texttt{IIA} on the evaluation set separately for each mechanism and language model, following \citet{wu2023interpretability}.
For mechanism \ModelS\ we use a stereotype-specific variant $\texttt{IIA}_\text{stereotype}$ that checks whether the post-intervention prediction matches the stereotypical pronoun for the source occupation rather than the source sentence's explicit pronoun.
Further details are provided in Appendix \ref{app:das-training}.

\section{DAS Recovers Three Subspaces}
\label{sec: Distributed-Alignment-Search}

We use boundless DAS to test whether the three hypothesized mechanisms are causally implemented.
First, we do a layer search to locate each mechanism (\S\ref{sec:layersearch}).
Then, we train fully on all diagnostic datasets (\S\ref{sec:full-training}).
Finally, we experiment with cross-mechanism transfer to evaluate whether the subspaces we recover are distinct (\S\ref{app:validation-das}).

\subsection{Layer Search}%
\label{sec:layersearch}

Since DAS intervenes on a single layer at a time, and different causal mechanisms may be encoded at different network depths, we search for the layer at which each mechanism is most strongly represented.
Table \ref{tab:layer-search-results} reports the best layer and corresponding IIA for each model and mechanism (full layer research results are shown in Appendix \ref{app:layer-search-all}).
We find that \textit{every mechanism is recoverable somewhere in the network};
no hypothesis can be ruled out on the grounds that its subspace does not exist.%

\paragraph{Different mechanisms dominate per model.}
Which mechanism is most recoverable, however, varies by model, and no single mechanism dominates.
\ModelG\ and \ModelR\ generally reach higher IIA than \ModelS: In Llama-3.1-8B, OLMo-2-1B, and Gemma-2-9B all three reach 0.78, 0.78, and 1.0, against 0.74, 0.60, and 0.90 for \ModelS\, and in OLMo-2-7B \ModelG\ reaches 0.935 versus 0.665 for \ModelS. The ordering is not universal, though, as in Qwen2.5-7B and OLMo-2-13B \ModelS\ leads instead (0.84 and 0.76), ahead of \ModelR\ (0.76, 0.72) and \ModelG\ (0.74, 0.62).

\begin{table}[t]
\centering
\small
\begin{tabularx}{\linewidth}{Xlrr}
\toprule
\textbf{Model} & \textbf{Mechanism} & \textbf{$\ell_{best}$} & \textbf{$IIA_{Best}$} \\
\midrule
\multirow{3}{*}{Llama-8B}
  & Group Binding \ModelG\ & $8$  & $\mathbf{0.78}$ \\
  & Stereotype \ModelS\    & $8$  & $0.74$ \\
  & Recency \ModelR\       & $8$  & $\mathbf{0.78}$ \\
\midrule
\multirow{3}{*}{OLMo-1B}
  & Group Binding \ModelG\ & $12$ & $\mathbf{0.78}$ \\
  & Stereotype \ModelS\    & $2$  & $0.60$ \\
  & Recency \ModelR\       & $6$  & $\mathbf{0.78}$ \\
\midrule
\multirow{3}{*}{OLMo-7B}
  & Group Binding \ModelG\ & $14$ & $\mathbf{0.94}$ \\
  & Stereotype \ModelS\    & $14$ & $0.67$ \\
  & Recency \ModelR\       & $16$ & $0.64$ \\
\midrule
\multirow{3}{*}{OLMo-13B}
  & Group Binding \ModelG\ & $10$ & $0.62$ \\
  & Stereotype \ModelS\    & $10$ & $\mathbf{0.76}$ \\
  & Recency \ModelR\       & $5$  & $0.72$ \\
\midrule
\multirow{3}{*}{Gemma-9B}
  & Group Binding \ModelG\ & $10$ & $\mathbf{1.00}$ \\
  & Stereotype \ModelS\    & $15$ & $0.90$ \\
  & Recency \ModelR\       & $10$ & $\mathbf{1.00}$ \\
\midrule
\multirow{3}{*}{Qwen-7B}
  & Group Binding \ModelG\ & $14$ & $0.74$ \\
  & Stereotype \ModelS\    & $7$  & $\mathbf{0.84}$ \\
  & Recency \ModelR\       & $14$ & $0.76$ \\
\bottomrule
\end{tabularx}
\caption{Best layer and IIA per model and mechanism.}
\label{tab:layer-search-results}
\end{table}

\paragraph{The mechanisms sit at different depths.}
Within a single model the best layer usually differs across mechanisms; in OLMo-2-7B, for instance, \ModelG\ sits at layer 8, \ModelR\ at layer 6, and \ModelS\ at layer 0.
This is a first indication that we are recovering three separate subspaces rather than one shared representation.
However, this evidence is only suggestive; since IIA here is measured on the mixed dataset, two mechanisms can score similarly high simply because they happen to agree on a prediction.
We disentangle this in the experiments that follow.

\subsection{Full Training}
\label{sec:full-training}

\begin{table*}[t]
\centering
\small
\begin{tabularx}{\linewidth}{Xlrrrrrr}
\toprule
\textbf{Dataset} & \textbf{Mechanism}
  & \textbf{Llama-8B}
  & \textbf{OLMo-1B}
  & \textbf{OLMo-7B}
  & \textbf{OLMo-13B}
  & \textbf{Gemma-9B}
  & \textbf{Qwen-7B} \\
\midrule
\multirow{3}{*}{\texttt{GR} (isolates \ModelS{})}
  & Group Binding \ModelG & $0.59$ & $0.25$ & $0.49$ & $0.66$ & $0.60$ & $0.92$ \\
  & Stereotype \ModelS & $0.58$ & $0.38$ & $\mathbf{0.69}$ & $\mathbf{0.89}$ & $0.60$ & $\mathbf{0.95}$ \\
  & Recency \ModelR & $\mathbf{0.68}$ & $\mathbf{0.85}$ & $0.60$ & $0.735$ & $\mathbf{1.00}$ & $0.82$ \\
\midrule
\multirow{3}{*}{\texttt{GS} (isolates \ModelR{})}
  & Group Binding \ModelG & $0.58$ & $0.40$ & $0.56$ & $0.89$ & $0.55$ & $0.96$ \\
  & Stereotype \ModelS & $0.59$ & $0.47$ & $0.69$ & $\mathbf{0.89}$ & $0.68$ & $\mathbf{0.97}$ \\
  & Recency \ModelR & $\mathbf{0.80}$ & $\mathbf{0.89}$ & $\mathbf{0.84}$ & $0.78$ & $\mathbf{0.70}$ & $0.87$ \\
\midrule
\multirow{3}{*}{\texttt{RS} (isolates \ModelG{})}
  & Group Binding \ModelG & $\mathbf{0.75}$ & $\mathbf{0.79}$ & $\mathbf{0.75}$ & $0.76$ & $0.52$ & $0.93$ \\
  & Stereotype \ModelS & $0.71$ & $0.56$ & $0.53$ & $\mathbf{0.77}$ & $\mathbf{0.55}$ & $\mathbf{1.00}$ \\
  & Recency \ModelR & $0.69$ & $0.59$ & $0.74$ & $0.63$ & $\mathbf{0.55}$ & $0.95$ \\
\bottomrule
\end{tabularx}
\caption{Best IIA per mechanism and dataset across models after full training on the diagnostic datasets.}
\label{tab:das-results-full}
\end{table*}

Having located each mechanism, we now train DAS at the selected layers on the full diagnostic datasets, each of which isolates one mechanism by construction (see \S\ref{sec:data}).
Table \ref{tab:das-results-full} reports the results.

\paragraph{All three mechanisms remain causally active.}
In most cases, the mechanism that scores highest in the layer search remains strong across the three diagnostic datasets, though this is not consistent for every model.
\ModelS\ is the strongest mechanism for Qwen2.5-7B as before, and now also for Gemma2-9B, despite having been its weakest mechanism during layer search.
For Llama, OLMo-1B, OLMo-7B the dominant mechanism also carries over, with \ModelR\ performing best on \texttt{GR} and \texttt{GS} whereas \ModelG\ dominates on the \texttt{RS} dataset.
Across all datasets, no single mechanism dominates, and the IIAs are broadly comparable.
This reinforces the idea that pronoun fidelity is not a single-mechanism problem. 

\begin{table}[t]
\centering
\small
\begin{tabularx}{\linewidth}{Xrrrr}
\toprule
\textbf{Model} & \textbf{Rotation matrix} $\mathbf{R}$ & \texttt{GS} & \texttt{RS} & \texttt{GR} \\
\midrule
\multirow{3}{*}{Llama-8B}
  & \ModelG\ ($\ell = 8$)  & $0.55$ & $0.55$ & $\mathbf{0.65}$ \\
  & \ModelS\ ($\ell = 8$)  & $0.59$ & $\mathbf{0.60}$ & $0.60$ \\
  & \ModelR\ ($\ell = 8$)  & $\mathbf{0.68}$ & $0.49$ & $0.32$ \\
\midrule
\multirow{3}{*}{OLMo-1B}
  & \ModelG\ ($\ell = 12$) & $0.47$ & $0.36$ & $0.33$ \\
  & \ModelS\ ($\ell = 6$)  & $0.63$ & $0.50$ & $\mathbf{0.40}$ \\
  & \ModelR\ ($\ell = 2$)  & $\mathbf{0.89}$ & $\mathbf{0.83}$ & $0.13$ \\
\midrule
\multirow{3}{*}{OLMo-7B}
  & \ModelG\ ($\ell = 8$)  & $0.35$ & $0.30$ & $\mathbf{0.63}$ \\
  & \ModelS\ ($\ell = 6$)  & $0.34$ & $0.33$ & $0.68$ \\
  & \ModelR\ ($\ell = 0$)  & $\mathbf{0.74}$ & $\mathbf{0.53}$ & $0.57$ \\
\midrule
\multirow{3}{*}{OLMo-13B}
  & \ModelG\ ($\ell = 10$) & $0.54$ & $0.56$ & $0.80$ \\
  & \ModelS\ ($\ell = 10$) & $0.54$ & $0.61$ & $\mathbf{0.82}$ \\
  & \ModelR\ ($\ell = 0$)  & $\mathbf{0.73}$ & $\mathbf{0.75}$ & $0.58$ \\
\midrule
\multirow{3}{*}{Gemma-9B}
  & \ModelG\ ($\ell = 10$) & $0.26$ & $0.51$ & $0.44$ \\
  & \ModelS\ ($\ell = 10$) & $0.14$ & $\mathbf{0.70}$ & $\mathbf{0.70}$ \\
  & \ModelR\ ($\ell = 10$) & $\mathbf{0.99}$ & $0.12$ & $0.00$ \\
\midrule
\multirow{3}{*}{Qwen-7B}
  & \ModelG\ ($\ell = 14$) & $0.32$ & $0.34$ & $\mathbf{0.94}$ \\
  & \ModelS\ ($\ell = 7$)  & $0.39$ & $0.37$ & $0.81$ \\
  & \ModelR\ ($\ell = 14$) & $\mathbf{0.90}$ & $\mathbf{0.78}$ & $0.53$ \\
\bottomrule
\end{tabularx}
\caption{Cross-mechanism specificity per model. Rows index the trained rotation matrix and columns the diagnostic dataset. The best rotation per dataset is \textbf{bolded}. \ModelG\ and \ModelS\ tend to peak together on \texttt{GR} in most models, while \ModelR\ peaks on \texttt{GS}, showing that \ModelG\ and \ModelS\ behave alike while \ModelR\ is distinct.}
\label{tab:transfer-all}
\end{table}

\subsection{Validation}
\label{app:validation-das}
The layer search and full training establish that each mechanism is individually sufficient to flip a prediction, but not that the three subspaces are genuinely distinct.
We test specificity using the structure of the diagnostic datasets (\S\ref{sec:data}), since each isolates one mechanism.
We take the three rotation matrices $R_G$, $R_R$, $R_S$,
freeze them, and evaluate each on every dataset, giving a $3\times3$ matrix per model (rotation $\times$ evaluation dataset; Table \ref{tab:transfer-all}).

\paragraph{Recency is distinct; group binding and stereotype overlap.}
Across all models, \ModelR\ is the most distinct rotation:
It dominates \texttt{GS} and transfers beyond this to \texttt{RS} as well.
With \texttt{GR}, in contrast, either \ModelG\ or \ModelS wins, and the two are never clearly distinct from each other on any dataset.
In general, 
the \ModelG\ and \ModelS\ rotations tend to score similarly high or similarly low on the same datasets, whereas \ModelR\ scores high exactly where the others score low, and vice versa.
\ModelG\ and \ModelS\ thus behave as one overlapping subspace, while \ModelR\ stays separable.

\paragraph{Takeaway.}
We recover three causal subspaces for pronoun fidelity, each of which is sufficient to flip predictions, without a single mechanism dominating.
Cross-mechanism transfer shows that \ModelG\ and \ModelS\ behave similarly, while \ModelR\ stays distinct.

\section{A Mixture of Mechanisms Explains Model Behaviour}

\newcommand{\worst}[1]{\textcolor{black!45}{#1}}

\begin{table*}[t]
\centering
\small
\begin{tabularx}{\linewidth}{Xr|r|rrrrrr}
\toprule
\textbf{Model} 
  & $\mathbf{\ell_G / \ell_R / \ell_S}$
  & $\mathbf{\mathcal{M}}$
  & $\mathcal{M}\!\setminus\!\{G\}$
  & $\mathcal{M}\!\setminus\!\{R\}$
  & $\mathcal{M}\!\setminus\!\{S\}$
  & $\mathcal{M}\!\setminus\!\{G,R\}$
  & $\mathcal{M}\!\setminus\!\{G,S\}$
  & $\mathcal{M}\!\setminus\!\{R,S\}$\\
\midrule
Llama-8B
  & $8 / 8 / 8$
  & $0.970$
  & $\mathbf{0.974}$
  & $0.972$
  & $0.967$
  & $\worst{0.932}$
  & $0.966$
  & $0.949$ \\

Llama-70B
  & $48 / 0 / 48$
  & $\mathbf{0.943}$
  & $0.934$
  & $0.920$
  & $0.936$
  & $0.919$
  & $0.918$
  & $\worst{0.889}$ \\

OLMo-1B
  & $12 / 2 / 6$
  & $\mathbf{0.995}$
  & $0.994$
  & $0.991$
  & $0.983$
  & $0.991$
  & $0.960$
  & $\worst{0.908}$ \\

OLMo-7B
  & $14 / 14 / 16$
  & $\mathbf{0.961}$
  & $\mathbf{0.961}$
  & $0.943$
  & $0.945$
  & $0.943$
  & $0.943$
  & $\worst{0.884}$ \\

OLMo-13B
  & $10 / 10 / 5$
  & $0.953$
  & $\mathbf{0.975}$
  & $\worst{0.928}$
  & $0.931$
  & $\worst{0.928}$
  & $0.946$
  & $0.959$ \\

OLMo-32B
  & $60 / 60 / 60$
  & $\mathbf{0.991}$
  & $0.987$
  & $0.988$
  & $0.987$
  & $0.984$
  & $0.985$
  & $\worst{0.980}$ \\

Gemma-9B
  & $10 / 15 / 10$
  & $\mathbf{0.958}$
  & $0.957$
  & $0.947$
  & $0.935$
  & $0.946$
  & $0.933$
  & $\worst{0.874}$ \\

Gemma-27B
  & $40 / 40 / 0$
  & $\mathbf{0.976}$
  & $0.952$
  & $0.962$
  & $0.954$
  & $0.915$
  & $0.926$
  & $\worst{0.898}$ \\

Qwen-7B
  & $14 / 7 / 14$
  & $\mathbf{0.915}$
  & $\mathbf{0.915}$
  & $0.871$
  & $0.859$
  & $0.870$
  & $0.856$
  & $\worst{0.700}$ \\

Qwen-72B
  & $48 / 0 / 0$
  & $\mathbf{0.964}$
  & $0.962$
  & $0.947$
  & $0.936$
  & $0.947$
  & $0.935$
  & $\worst{0.864}$ \\
  
\bottomrule
\end{tabularx}
\caption{Jensen-Shannon similarity between predicted and empirical pronoun distributions for the full mixture model $\mathcal{M}$ and selected ablations. Higher is better. The best value in each row is shown in bold; the lowest value is shown in gray. $\ell_G$, $\ell_R$, and $\ell_S$ denote the layer selected for each mechanism.}
\label{tab:mixture_results}
\end{table*}

If no single mechanism dominates, \textit{how do the three combine to produce model behaviour?}
Following the causal model combination approach of \citet{pmlr-v275-pislar25a} and \citet{gur2025mixing}, we fit a lightweight mixture model on the mechanisms.

\paragraph{Fitting the mixture model.}
For each mechanism we collect empirical next-token distributions over \texttt{\{he, she, they\}} via inference-only interchange interventions:
We patch the source residual stream into the base run at the best layer $\ell$ identified in our DAS layer search (\S\ref{sec:layersearch}), with no trained rotation, and read off the softmax over the three pronouns.
For larger models where a full DAS training layer search is computationally infeasible (32B–72B), we instead identify $\ell$ with an inference-only search over a 200-example subset of the mixed datasets, applying \texttt{VanillaIntervention} patches \citep{wu-etal-2024-pyvene} that overwrite the base residual stream with the source vector. We select the last layer at which the base pronoun still dominates, i.e., the point just before the intervention takes effect and the model commits to an answer.

Each softmax distribution is indexed at the response position by the triple $(g,r,s)$: The pronouns predicted by group binding, recency, and stereotype, respectively. The mixture model predicts: 
\begin{equation}
\begin{aligned}
\text{logit}(a) ={}& w_{\mathcal{G}} \cdot \mathbf{1}\{a = g\} \\
&+ w_{\mathcal{R}}[r] \cdot \mathbf{1}\{a = r\} \\
&+ w_{\mathcal{S}}[s] \cdot \mathbf{1}\{a = s\},
\end{aligned}
\end{equation}
where $a\in\{he, she, they\}$. The scalar $w_{\mathcal{G}}$ and the pronoun-indexed vectors $w_{\mathcal{R}}[\cdot]$ and $w_{\mathcal{S}}[\cdot]$ are trained with Adam (learning rate $0.05$) and a Jensen-Shannon divergence loss. 
We report Jensen-Shannon similarities ($\text{JSS}\in [0,1]$, $1-JSD$) in Table \ref{tab:mixture_results} for the mixture model $\mathcal{M}$ and all single- and double-mechanism ablations.
Higher values show agreement between the mixture model's prediction and the empirical distribution of model behaviour.
The full model consistently achieves high JSS ($0.915$--$0.995$), indicating that the three mechanisms together account for almost all behaviour.

\paragraph{Recency and stereotype carry most of the signal.} Ablations where recency and stereotype are jointly removed ($\mathcal{M}\!\setminus\!\{R,S\}$) causes JSS to drop the most across nearly all models,
while removing either alone causes smaller but consistent degradation. %
We can thus conclude that model behaviour relies primarily on these two heuristics.

\paragraph{Group binding contributes little.}
Removing group binding ($\mathcal{M}\setminus\{G\}$) sometimes leaves JSS unchanged, or even improves it, matching or exceeding the full model across several models (by $0.022$ in OLMo-13B).
Rather than adding a signal beyond what the recency and stereotype mechanisms already capture, \textit{group binding can add noise.}

These patterns even line up somewhat with model-specific DAS results from \S\ref{sec: Distributed-Alignment-Search}:
In Qwen-7B, where \ModelS\ is the dominant subspace, ablating $\mathcal{M}\setminus\{S\}$ hurts JSS, dropping from $0.915$ to $0.859$;
in OLMo-13B, where \ModelG\ achieves lowest IIA, removing it from the mixture model actually helps. 

\paragraph{Takeaway.}
Pronoun fidelity can be fully explained by a combination of mechanisms: Recency and stereotype carry most of the signal, while group binding contributes little and can even add noise.

\section{Attention Head Analysis}
\label{sec:attention-heads}

So far we have shown
which mechanisms are at play in pronoun fidelity, where they are located, and how they combine,
but not which components of a model \textit{implement} them.
Inspired by \citet{feucht2025dual}, we look at individual attention heads.

\citet{feucht2025dual} find that LLMs use attention heads to implement two parallel copying routes: \textit{Token-level induction heads} match exact individual subword tokens, while \textit{concept-level induction heads} attend to the ends of multi-token words and copy full words as abstract concepts for tasks like translations and paraphrasing.
In our setting, we posit that concept-level copying corresponds to retrieving a bound semantic unit (the occupation-pronoun pair) which is what both group binding \ModelG\ and stereotype bias \ModelS\ rely on.
Token-level copying corresponds to recency \ModelR, where the model repeats the most recent pronoun.
To test whether concept-level or token-level copying dominates pronoun fidelity, we identify and ablate the attention heads responsible for each copying route.

\paragraph{Identification.}
\begin{table*}[t]
\centering
\small
\begin{tabularx}{\linewidth}{Xp{6.3cm}p{6.3cm}}
\toprule
\textbf{Dataset} & \textbf{Clean - Example} & \textbf{Corrupt - Example} \\
\midrule
\texttt{concept-level} (pron: \textit{he})
& The technician thought that \textit{he} had been having a bad day. The customer felt energetic; \textit{she} had slept deeply. The technician said that \_\_\_ had completed the repair.
& The customer felt energetic; \textit{she} had slept deeply. The technician thought that \textit{he} had been having a bad day. The technician said that \_\_\_ had completed the repair.\\[0.75ex]
\hdashline
\\[-0.25ex]
\texttt{token-level}
(pron: \textit{she})
& The customer felt energetic; \textit{she} had slept deeply. The technician thought that \textit{he} had been having a bad day. The technician said that \_\_\_ had completed the repair.
& The technician thought that \textit{he} had been having a bad day. The customer felt energetic; \textit{she} had slept deeply. The technician said that \_\_\_ had completed the repair.\\
\bottomrule
\end{tabularx}
\caption{Examples from the two copying datasets. Each dataset contains 50 data points to identify the heads and 50 data points on which we ablate. Clean activations are patched into the corrupt run.}
\label{tab:attention-head-ds}
\end{table*}

We identify attention heads responsible for a particular copying route by patching pre-projection outputs from clean to corrupt instances of two new datasets that isolate concept-level and token-level copying (see Table \ref{tab:attention-head-ds}).
If the head carries information about the occupation-pronoun group as a concept, then it will ignore recency information and consistently predict \textit{he} on clean and corrupt instances of the \texttt{concept-level} dataset.
In contrast, the \texttt{token-level} dataset tests whether the head directly copies the last seen pronoun token (e.g., copying \textit{he} in the clean version and \textit{she} in the corrupt version of the example).

For each attention head $a^\texttt{(l,h)}$, we patch its pre-projection output at the final token position from the clean run into the corrupt run, and measure
\[
\Delta_{\text{log P}}^{(l,h)} = \log P_{\text{patched}}(\text{pron}) - \log P_{\text{corrupt}}(\text{pron}),
\]
where ``pron'' is the target entity's pronoun in the concept copying experiment, and the distractor pronoun in the token-level experiment.
Heads with high $\Delta_{\text{log P}}^{(l,h)}$ in the concept copying dataset are concept induction heads;
heads with high $\Delta_{\text{log P}}^{(l,h)}$ on the token-level dataset implement token-level copying.

\paragraph{Ablation.}

\begin{table}[t]
\centering
\small
\begin{tabularx}{\linewidth}{Xrrr}
\toprule
\textbf{Model} & \textbf{Heads}
& $\mathbf{\Delta A} (A_{abl} - A_{base})$
& $\mathbf{\Delta A_{rd}}$ \\
\midrule
\rowcolor{grey} \multicolumn{4}{c}{\texttt{concept-level}} \\
Llama-8B   & 11 & $-0.04$ ($0.96 - 1.00$) & $-0.04$ \\
OLMo-1B      & 11 & $-0.78$ ($0.16 - 0.94$) & $-0.07$ \\
OLMo-7B      & 11 & $-0.64$ ($0.16 - 0.80$) & $0.00$ \\
OLMo-13B     & 16 & $-0.28$ ($0.70 - 0.98$) & $0.00$ \\
Gemma-9B     & 23 & $-0.12$ ($0.86 - 0.98$) & $-0.02$ \\
Qwen-7B     & 29 & $-0.58$ ($0.30 - 0.88$) & $-0.07$ \\
\midrule
\rowcolor{grey} \multicolumn{4}{c}{\texttt{token-level}} \\
Llama-8B & 21 & $+0.40$ ($0.76 - 0.48$) & $-0.04$ \\
OLMo-1B    & 72 & $-0.36$ ($0.44 - 0.80$) & $-0.22$ \\
OLMo-7B    & 41 & $-0.36$ ($0.08-0.34$) & $-0.01$ \\
OLMo-13B   & 61 & $-0.04$ ($0.64 - 0.68$) & $0.00$ \\
Gemma-9B   & 49 & $-0.16$ ($0.64 - 0.80$) & $-0.03$ \\
Qwen-7B   & 1  & $+0.44$ ($0.90 - 0.46$) & $-0.05$ \\
\bottomrule
\end{tabularx}
\caption{Accuracy with cumulative head ablations for concept-level and token-level copying. $A_{base}$/$A_{abl}$ denote base and ablated accuracy; $A_{rd}$ shows accuracy when ablating the same number of heads at random.}
\label{tab:ablation-results-combined}
\end{table}

We ablate heads by zeroing out the pre-projection $a^{(l,h)}$ of each top-k (k=$100$) head at the final token position of the clean prompt.
We then measure the drop in $\log P(\text{pron})$: 
\[
    \Delta_{\text{ablate}}^{(l,h)} = \log P_{\text{ablated}}(\text{pron}) - \log P_{\text{clean}}(\text{pron}).
\]

\noindent We perform \textit{single head ablations} with one top-k head zeroed out at a time, and \textit{cumulative ablations}, where heads are zeroed in rank order.
Both use $N=50$ held-out examples not seen during identification, scored by log probabilities.

\paragraph{Results.}

We show accuracy drops with cumulative head ablation in Table \ref{tab:ablation-results-combined}.
Further results are shown in Appendix \ref{app:attention-heads}.
For \textbf{concept-level copying}, a small set of heads have substantial causal effects:
Removing identified heads causes large, consistent accuracy drops.
OLMo-1B and OLMo-7B both fall to $0.16$ after ablating only 11 heads; Qwen-7B drops to $0.30$ after 29.
A random ablation baseline, removing the same number of heads at random, has almost no effect,
confirming the drop is specific to the identified heads.
This provides evidence of a localizable concept-copying route.
Llama-3.1-8B is the exception: its targeted drop matches its random baseline, consistent with its flat causal scores and a more distributed concept route.

For \textbf{token-level copying},
there is no distinct set of high-scoring heads, suggesting the mechanism is distributed across many heads.
Many heads must be ablated for accuracy to change, e.g., OLMo-2-1B degrades from 0.80 to 0.44 only after ablating 72 heads, and OLMo-13B is almost unaffected after ablating 61 heads.
Interestingly, for Qwen-7B and Llama-8B, ablating the identified token-copying heads does not lower accuracy but \textit{raises} it ($+0.44$ and $+0.40$).
This indicates competition between the two copying routes during pronoun resolution under discourse complexity.

\paragraph{Takeaway.}
Pronoun fidelity is solved via two competing copying routes that are localized differently:
Concept copying through a few heads, and token copying distributed across many low-effect heads. 
They compete, as evidenced by ablating token copying heads flipping a model from recency-driven bias to correct resolution, but which route dominates is model-dependent.
These results also explain the cross-mechanism transfer results from \S\ref{app:validation-das}:
\ModelG\ and \ModelS\ share the concept-copying route and thus overlapping subspaces despite drawing on different information, while \ModelR\ takes the token-copying route and stays distinct.

\section{Related Works}

\paragraph{Pronoun Fidelity}
Faithful pronoun reuse was first studied in \citet{ovalle-et-al-2023-tango} and \citet{hossain-etal-2023-misgendered}, across a range of pronouns but in settings with limited discourse complexity.
\citet{gautam-etal-2024-robust} added discourse complexity to the task, using sentences about additional referents in their RUFF dataset, which we use.
These datasets have been used in a comparison of probability- and generation-based evaluations in \citep{subramonian2025agree}, as well as in an evaluation of reasoning models in \citet{gautam-2026-training} (which we do not consider).
More recently, \citet{kotek2026protextbenchmarkdatasetmeasuring} introduce ProText, which contains natural narratives about people, but still focuses on single rather than multiple discourse entities.
All these works take a behavioural perspective on pronoun fidelity.

\paragraph{Mechanistic Interpretability} 
Mechanistic interpretability is a framework for discovering causal mechanisms that explain how inputs are transformed into outputs via intermediate representations \citep{mueller2024missed,saphra2024mechanistic}.
This is complicated by \textit{superposition} in neural networks, i.e., concepts are distributed across neurons, and a single neuron might encode multiple concepts \cite{elhage2022toy}.
Distributed Alignment Search \citep[DAS;][]{wu2023interpretability,pmlr-v236-geiger24a} addresses this by searching for linearly encoded causal subspaces in the residual stream, grounded in causal abstraction theory \cite{geiger2021causal, geiger2025causal}.
Supervised DAS is the strongest method for causal variable localization according to a benchmark of popular mechanistic interpretability methods \citep{mueller2025mib}, motivating our use of this method.
DAS has also been applied to interpret how LLMs process filler-gap dependencies in English \citep{boguraev2025causal}.

\paragraph{Interpreting Pronoun Fidelity}
Although no paper directly tackles pronoun fidelity,
prior work on stereotypical biases and entity tracking are relevant.

\citet{vig2020investigating} introduce Causal Mediation Analysis (also known as activation patching) 
isolate small sets of neurons and attention heads responsible for gender bias.
\citet{chintam-etal-2023-identifying} extend this
to localize bias-related components in GPT-2.
\citet{bashir2025dissecting} use Edge Attribution Patching to discover bias-related circuits, but find them highly unstable, and \citet{hanna2024have} provide evidence for fundamental limitations of circuit discovery methods.
These works suggest that stereotypical associations are not cleanly localized and affect other tasks, motivating our study of their interaction with other mechanisms.

Entity tracking, a prerequisite to long-context understanding and coherent generation \cite{kim2023entity}, is also implicated in pronoun fidelity.
\citet{prakash2024fine} use activation patching and circuit discovery to show how collaborating attention heads track entities via positions.
However, \citet{gur2025mixing} find that a purely positional mechanism generalizes poorly with more bound entities.
They propose a causal model combining multiple mechanisms, motivating our approach.

\section{Conclusion}
We present the first mechanistic study of pronoun fidelity, asking whether failure to track pronouns faithfully is driven by reasoning, repetition, or stereotypical bias. 
We find that it is driven by a combination of all three.
Boundless DAS recovers group binding \ModelG, recency \ModelR, and stereotype \ModelS\ as three separable causal subspaces, and a mixture of them predicts 91-99.5\% of model behaviour. Most models lean on recency and stereotype bias, while group binding contributes little and occasionally adds noise.
These mechanisms appear to be implemented through attention heads that carry pronoun information via two competing copying routes.
One route retrieves a semantic unit through a few localized heads; the other repeats surface forms through many distributed heads, and which route is taken is model-dependent. 
Group binding and stereotype share the concept-copying route, differing only in whether the retrieved unit is a context-bound entity or stereotypical bias, while recency uses the token-copying route. 
In summary, pronoun fidelity failure is not a single-mechanism problem.
It arises from competition among mechanisms that are all active at once, with the outcome determined by which mechanism wins.

\section*{Limitations}

\paragraph{Inherited limitations of the RUFF dataset.}
RUFF is a synthetically-created template-based dataset, which sacrifices ecological validity for tight control over the entities and pronouns involved in each data instance.
This control makes the dataset natural for use with Distributed Alignment Search, which requires tightly-paired variants to isolate mechanisms, but it may not be representative of how reference ``in the wild'' works, cf. OntoNotes \citep{Weischedel-et-al-ontonotes} or the GUM corpus \citep{Zeldes_2017}.
Additionally, RUFF is an English-only dataset.
Morphologically richer languages (e.g., languages that mark gender on other parts of speech, or that mark more cases, etc.), might have entirely different mechanisms.

\paragraph{Limitations of our use of RUFF.}
Although RUFF contains data instances with 0-5 distractor sentences, we only include the 0 and 1 distractor settings, as these represent the most dramatic performance gap.
Additional distractor sentences may modulate the effects of the different mechanisms.
In our restricted setting, we also only experiment with 3 English pronouns (he, she, they) that are tokenized singly.
Unlike \citet{gautam-etal-2024-robust}, we do not consider neopronouns as they are usually tokenized into multiple tokens \citep{ovalle-etal-2024-tokenization}, and therefore cannot say whether the mechanisms for neopronoun fidelity are the same.

\paragraph{Restrictions from Distributed Alignment Search.}
As we use the three available singular pronouns in English that can be singly tokenized, it is not possible to have a single diagnostic dataset for DAS where all mechanisms can be represented, as one pronoun always represents the stereotype mechanism, one represents the group binding mechanism, and the other the repetition mechanism.
Additionally, it is possible that there is another single mechanism beyond the ones we cover that would explain model pronoun fidelity as well as our combined mixed-mechanism explanation, although it is likely to be less interpretable.

\paragraph{Additional MI methods and models.}
Our research questions can be answered with many other approaches within mechanistic interpretability, notably circuit discovery.
However, there is evidence that large neural language models may often implement the same task in multiple, distinct circuits \citep{chen2026circuitsleadromerethinking}, making them less robust as explanations of model mechanisms than the learned rotation matrices of boundless DAS.
In general, we take the position that different methodological approaches shed light on different aspects of model interpretability, which is why we also experiment with activation patching in our attention head analysis.
Our findings apply to the class of models we study, which is instruction-tuned decoder-only language models, but our work could be expanded to other models, e.g., encoder-only models, which show very different patterns on the RUFF dataset, and reasoning models, which make different patterns of mistakes \citep{gautam-2026-training} and may therefore implement different mechanisms for pronominal reasoning.

\section*{Ethics Statement}

Although we limit our study to the three most widely-used third-person singular pronouns in English that are therefore singly tokenized (\textit{he}, \textit{she}, and \textit{they}), we note that other neopronouns are also in use in English.
As with behavioural studies of model fairness, mechanistic studies like ours also come with the potential risk of misuse to perpetuate stereotypical narratives---in our case, pertaining to gender.
Additionally, applications involving tracking person entities through reference in natural language can be used for surveillance, which we stand against.
We use all models and the RUFF dataset in accordance with their licenses and terms of use.

\section*{Acknowledgments}
Vagrant Gautam's work is funded by the Klaus Tschira Foundation, Heidelberg, Germany. Jesujoba Alabi was funded by the Deutsche Forschungsgemeinschaft (DFG, German Research Foundation) – Project-ID 232722074 – SFB 1102.

\bibliography{custom}

@article{
bashir2025dissecting,
title={Dissecting Bias in {LLM}s: A Mechanistic Interpretability Perspective},
author={Zubair Bashir and Bhavik Chandna and Procheta Sen},
journal={Transactions on Machine Learning Research},
issn={2835-8856},
year={2025},
url={https://openreview.net/forum?id=EpQ2CBJTjD},
note={}
}

@article{gautam-etal-2024-robust,
    title = "Robust Pronoun Fidelity with {E}nglish {LLM}s: Are they Reasoning, Repeating, or Just Biased?",
    author = "Gautam, Vagrant  and
      Bingert, Eileen  and
      Zhu, Dawei  and
      Lauscher, Anne  and
      Klakow, Dietrich",
    journal = "Transactions of the Association for Computational Linguistics",
    volume = "12",
    year = "2024",
    address = "Cambridge, MA",
    publisher = "MIT Press",
    url = "https://aclanthology.org/2024.tacl-1.95/",
    doi = "10.1162/tacl_a_00719",
    pages = "1755--1779"
}

@inproceedings{geiger2021causal,
 author = {Geiger, Atticus and Lu, Hanson and Icard, Thomas and Potts, Christopher},
 booktitle = {Advances in Neural Information Processing Systems},
 editor = {M. Ranzato and A. Beygelzimer and Y. Dauphin and P.S. Liang and J. Wortman Vaughan},
 pages = {9574--9586},
 publisher = {Curran Associates, Inc.},
 title = {Causal Abstractions of Neural Networks},
 url = {https://proceedings.neurips.cc/paper_files/paper/2021/file/4f5c422f4d49a5a807eda27434231040-Paper.pdf},
 volume = {34},
 year = {2021}
}

@InProceedings{pmlr-v236-geiger24a,
  title = 	 {Finding Alignments Between Interpretable Causal Variables and Distributed Neural Representations},
  author =       {Geiger, Atticus and Wu, Zhengxuan and Potts, Christopher and Icard, Thomas and Goodman, Noah},
  booktitle = 	 {Proceedings of the Third Conference on Causal Learning and Reasoning},
  pages = 	 {160--187},
  year = 	 {2024},
  editor = 	 {Locatello, Francesco and Didelez, Vanessa},
  volume = 	 {236},
  series = 	 {Proceedings of Machine Learning Research},
  month = 	 {01--03 Apr},
  publisher =    {PMLR},
  url = 	 {https://proceedings.mlr.press/v236/geiger24a.html}
}

@inproceedings{
mueller2025mib,
title={{MIB}: A Mechanistic Interpretability Benchmark},
author={Aaron Mueller and Atticus Geiger and Sarah Wiegreffe and Dana Arad and Iv{\'a}n Arcuschin and Adam Belfki and Yik Siu Chan and Jaden Fried Fiotto-Kaufman and Tal Haklay and Michael Hanna and Jing Huang and Rohan Gupta and Yaniv Nikankin and Hadas Orgad and Nikhil Prakash and Anja Reusch and Aruna Sankaranarayanan and Shun Shao and Alessandro Stolfo and Martin Tutek and Amir Zur and David Bau and Yonatan Belinkov},
booktitle={Forty-second International Conference on Machine Learning},
year={2025},
url={https://openreview.net/forum?id=sSrOwve6vb}
}

@inproceedings{vig2020investigating,
 author = {Vig, Jesse and Gehrmann, Sebastian and Belinkov, Yonatan and Qian, Sharon and Nevo, Daniel and Singer, Yaron and Shieber, Stuart},
 booktitle = {Advances in Neural Information Processing Systems},
 editor = {H. Larochelle and M. Ranzato and R. Hadsell and M.F. Balcan and H. Lin},
 pages = {12388--12401},
 publisher = {Curran Associates, Inc.},
 title = {Investigating Gender Bias in Language Models Using Causal Mediation Analysis},
 url = {https://proceedings.neurips.cc/paper_files/paper/2020/file/92650b2e92217715fe312e6fa7b90d82-Paper.pdf},
 volume = {33},
 year = {2020}
}

@inproceedings{wu2023interpretability,
 author = {Wu, Zhengxuan and Geiger, Atticus and Icard, Thomas and Potts, Christopher and Goodman, Noah},
 booktitle = {Advances in Neural Information Processing Systems},
 editor = {A. Oh and T. Naumann and A. Globerson and K. Saenko and M. Hardt and S. Levine},
 pages = {78205--78226},
 publisher = {Curran Associates, Inc.},
 title = {Interpretability at Scale: Identifying Causal Mechanisms in Alpaca},
 url = {https://proceedings.neurips.cc/paper_files/paper/2023/file/f6a8b109d4d4fd64c75e94aaf85d9697-Paper-Conference.pdf},
 volume = {36},
 year = {2023}
}

@article{gallegos2024bias,
    title = "Bias and Fairness in Large Language Models: A Survey",
    author = "Gallegos, Isabel O.  and
      Rossi, Ryan A.  and
      Barrow, Joe  and
      Tanjim, Md Mehrab  and
      Kim, Sungchul  and
      Dernoncourt, Franck  and
      Yu, Tong  and
      Zhang, Ruiyi  and
      Ahmed, Nesreen K.",
    journal = "Computational Linguistics",
    volume = "50",
    number = "3",
    month = sep,
    year = "2024",
    address = "Cambridge, MA",
    publisher = "MIT Press",
    url = "https://aclanthology.org/2024.cl-3.8/",
    doi = "10.1162/coli_a_00524",
    pages = "1097--1179"
}

@misc{grattafiori2024llama,
      title={The Llama 3 Herd of Models}, 
      author={Aaron Grattafiori and Abhimanyu Dubey and Abhinav Jauhri and Abhinav Pandey and Abhishek Kadian and Ahmad Al-Dahle and Aiesha Letman and Akhil Mathur and Alan Schelten and Alex Vaughan and Amy Yang and Angela Fan and Anirudh Goyal and Anthony Hartshorn and Aobo Yang and Archi Mitra and Archie Sravankumar and Artem Korenev and Arthur Hinsvark and Arun Rao and Aston Zhang and Aurelien Rodriguez and Austen Gregerson and Ava Spataru and Baptiste Roziere and Bethany Biron and Binh Tang and Bobbie Chern and Charlotte Caucheteux and Chaya Nayak and Chloe Bi and Chris Marra and Chris McConnell and Christian Keller and Christophe Touret and Chunyang Wu and Corinne Wong and Cristian Canton Ferrer and Cyrus Nikolaidis and Damien Allonsius and Daniel Song and Danielle Pintz and Danny Livshits and Danny Wyatt and David Esiobu and Dhruv Choudhary and Dhruv Mahajan and Diego Garcia-Olano and Diego Perino and Dieuwke Hupkes and Egor Lakomkin and Ehab AlBadawy and Elina Lobanova and Emily Dinan and Eric Michael Smith and Filip Radenovic and Francisco Guzmán and Frank Zhang and Gabriel Synnaeve and Gabrielle Lee and Georgia Lewis Anderson and Govind Thattai and Graeme Nail and Gregoire Mialon and Guan Pang and Guillem Cucurell and Hailey Nguyen and Hannah Korevaar and Hu Xu and Hugo Touvron and Iliyan Zarov and Imanol Arrieta Ibarra and Isabel Kloumann and Ishan Misra and Ivan Evtimov and Jack Zhang and Jade Copet and Jaewon Lee and Jan Geffert and Jana Vranes and Jason Park and Jay Mahadeokar and Jeet Shah and Jelmer van der Linde and Jennifer Billock and Jenny Hong and Jenya Lee and Jeremy Fu and Jianfeng Chi and Jianyu Huang and Jiawen Liu and Jie Wang and Jiecao Yu and Joanna Bitton and Joe Spisak and Jongsoo Park and Joseph Rocca and Joshua Johnstun and Joshua Saxe and Junteng Jia and Kalyan Vasuden Alwala and Karthik Prasad and Kartikeya Upasani and Kate Plawiak and Ke Li and Kenneth Heafield and Kevin Stone and Khalid El-Arini and Krithika Iyer and Kshitiz Malik and Kuenley Chiu and Kunal Bhalla and Kushal Lakhotia and Lauren Rantala-Yeary and Laurens van der Maaten and Lawrence Chen and Liang Tan and Liz Jenkins and Louis Martin and Lovish Madaan and Lubo Malo and Lukas Blecher and Lukas Landzaat and Luke de Oliveira and Madeline Muzzi and Mahesh Pasupuleti and Mannat Singh and Manohar Paluri and Marcin Kardas and Maria Tsimpoukelli and Mathew Oldham and Mathieu Rita and Maya Pavlova and Melanie Kambadur and Mike Lewis and Min Si and Mitesh Kumar Singh and Mona Hassan and Naman Goyal and Narjes Torabi and Nikolay Bashlykov and Nikolay Bogoychev and Niladri Chatterji and Ning Zhang and Olivier Duchenne and Onur Çelebi and Patrick Alrassy and Pengchuan Zhang and Pengwei Li and Petar Vasic and Peter Weng and Prajjwal Bhargava and Pratik Dubal and Praveen Krishnan and Punit Singh Koura and Puxin Xu and Qing He and Qingxiao Dong and Ragavan Srinivasan and Raj Ganapathy and Ramon Calderer and Ricardo Silveira Cabral and Robert Stojnic and Roberta Raileanu and Rohan Maheswari and Rohit Girdhar and Rohit Patel and Romain Sauvestre and Ronnie Polidoro and Roshan Sumbaly and Ross Taylor and Ruan Silva and Rui Hou and Rui Wang and Saghar Hosseini and Sahana Chennabasappa and Sanjay Singh and Sean Bell and Seohyun Sonia Kim and Sergey Edunov and Shaoliang Nie and Sharan Narang and Sharath Raparthy and Sheng Shen and Shengye Wan and Shruti Bhosale and Shun Zhang and Simon Vandenhende and Soumya Batra and Spencer Whitman and Sten Sootla and Stephane Collot and Suchin Gururangan and Sydney Borodinsky and Tamar Herman and Tara Fowler and Tarek Sheasha and Thomas Georgiou and Thomas Scialom and Tobias Speckbacher and Todor Mihaylov and Tong Xiao and Ujjwal Karn and Vedanuj Goswami and Vibhor Gupta and Vignesh Ramanathan and Viktor Kerkez and Vincent Gonguet and Virginie Do and Vish Vogeti and Vítor Albiero and Vladan Petrovic and Weiwei Chu and Wenhan Xiong and Wenyin Fu and Whitney Meers and Xavier Martinet and Xiaodong Wang and Xiaofang Wang and Xiaoqing Ellen Tan and Xide Xia and Xinfeng Xie and Xuchao Jia and Xuewei Wang and Yaelle Goldschlag and Yashesh Gaur and Yasmine Babaei and Yi Wen and Yiwen Song and Yuchen Zhang and Yue Li and Yuning Mao and Zacharie Delpierre Coudert and Zheng Yan and Zhengxing Chen and Zoe Papakipos and Aaditya Singh and Aayushi Srivastava and Abha Jain and Adam Kelsey and Adam Shajnfeld and Adithya Gangidi and Adolfo Victoria and Ahuva Goldstand and Ajay Menon and Ajay Sharma and Alex Boesenberg and Alexei Baevski and Allie Feinstein and Amanda Kallet and Amit Sangani and Amos Teo and Anam Yunus and Andrei Lupu and Andres Alvarado and Andrew Caples and Andrew Gu and Andrew Ho and Andrew Poulton and Andrew Ryan and Ankit Ramchandani and Annie Dong and Annie Franco and Anuj Goyal and Aparajita Saraf and Arkabandhu Chowdhury and Ashley Gabriel and Ashwin Bharambe and Assaf Eisenman and Azadeh Yazdan and Beau James and Ben Maurer and Benjamin Leonhardi and Bernie Huang and Beth Loyd and Beto De Paola and Bhargavi Paranjape and Bing Liu and Bo Wu and Boyu Ni and Braden Hancock and Bram Wasti and Brandon Spence and Brani Stojkovic and Brian Gamido and Britt Montalvo and Carl Parker and Carly Burton and Catalina Mejia and Ce Liu and Changhan Wang and Changkyu Kim and Chao Zhou and Chester Hu and Ching-Hsiang Chu and Chris Cai and Chris Tindal and Christoph Feichtenhofer and Cynthia Gao and Damon Civin and Dana Beaty and Daniel Kreymer and Daniel Li and David Adkins and David Xu and Davide Testuggine and Delia David and Devi Parikh and Diana Liskovich and Didem Foss and Dingkang Wang and Duc Le and Dustin Holland and Edward Dowling and Eissa Jamil and Elaine Montgomery and Eleonora Presani and Emily Hahn and Emily Wood and Eric-Tuan Le and Erik Brinkman and Esteban Arcaute and Evan Dunbar and Evan Smothers and Fei Sun and Felix Kreuk and Feng Tian and Filippos Kokkinos and Firat Ozgenel and Francesco Caggioni and Frank Kanayet and Frank Seide and Gabriela Medina Florez and Gabriella Schwarz and Gada Badeer and Georgia Swee and Gil Halpern and Grant Herman and Grigory Sizov and Guangyi and Zhang and Guna Lakshminarayanan and Hakan Inan and Hamid Shojanazeri and Han Zou and Hannah Wang and Hanwen Zha and Haroun Habeeb and Harrison Rudolph and Helen Suk and Henry Aspegren and Hunter Goldman and Hongyuan Zhan and Ibrahim Damlaj and Igor Molybog and Igor Tufanov and Ilias Leontiadis and Irina-Elena Veliche and Itai Gat and Jake Weissman and James Geboski and James Kohli and Janice Lam and Japhet Asher and Jean-Baptiste Gaya and Jeff Marcus and Jeff Tang and Jennifer Chan and Jenny Zhen and Jeremy Reizenstein and Jeremy Teboul and Jessica Zhong and Jian Jin and Jingyi Yang and Joe Cummings and Jon Carvill and Jon Shepard and Jonathan McPhie and Jonathan Torres and Josh Ginsburg and Junjie Wang and Kai Wu and Kam Hou U and Karan Saxena and Kartikay Khandelwal and Katayoun Zand and Kathy Matosich and Kaushik Veeraraghavan and Kelly Michelena and Keqian Li and Kiran Jagadeesh and Kun Huang and Kunal Chawla and Kyle Huang and Lailin Chen and Lakshya Garg and Lavender A and Leandro Silva and Lee Bell and Lei Zhang and Liangpeng Guo and Licheng Yu and Liron Moshkovich and Luca Wehrstedt and Madian Khabsa and Manav Avalani and Manish Bhatt and Martynas Mankus and Matan Hasson and Matthew Lennie and Matthias Reso and Maxim Groshev and Maxim Naumov and Maya Lathi and Meghan Keneally and Miao Liu and Michael L. Seltzer and Michal Valko and Michelle Restrepo and Mihir Patel and Mik Vyatskov and Mikayel Samvelyan and Mike Clark and Mike Macey and Mike Wang and Miquel Jubert Hermoso and Mo Metanat and Mohammad Rastegari and Munish Bansal and Nandhini Santhanam and Natascha Parks and Natasha White and Navyata Bawa and Nayan Singhal and Nick Egebo and Nicolas Usunier and Nikhil Mehta and Nikolay Pavlovich Laptev and Ning Dong and Norman Cheng and Oleg Chernoguz and Olivia Hart and Omkar Salpekar and Ozlem Kalinli and Parkin Kent and Parth Parekh and Paul Saab and Pavan Balaji and Pedro Rittner and Philip Bontrager and Pierre Roux and Piotr Dollar and Polina Zvyagina and Prashant Ratanchandani and Pritish Yuvraj and Qian Liang and Rachad Alao and Rachel Rodriguez and Rafi Ayub and Raghotham Murthy and Raghu Nayani and Rahul Mitra and Rangaprabhu Parthasarathy and Raymond Li and Rebekkah Hogan and Robin Battey and Rocky Wang and Russ Howes and Ruty Rinott and Sachin Mehta and Sachin Siby and Sai Jayesh Bondu and Samyak Datta and Sara Chugh and Sara Hunt and Sargun Dhillon and Sasha Sidorov and Satadru Pan and Saurabh Mahajan and Saurabh Verma and Seiji Yamamoto and Sharadh Ramaswamy and Shaun Lindsay and Shaun Lindsay and Sheng Feng and Shenghao Lin and Shengxin Cindy Zha and Shishir Patil and Shiva Shankar and Shuqiang Zhang and Shuqiang Zhang and Sinong Wang and Sneha Agarwal and Soji Sajuyigbe and Soumith Chintala and Stephanie Max and Stephen Chen and Steve Kehoe and Steve Satterfield and Sudarshan Govindaprasad and Sumit Gupta and Summer Deng and Sungmin Cho and Sunny Virk and Suraj Subramanian and Sy Choudhury and Sydney Goldman and Tal Remez and Tamar Glaser and Tamara Best and Thilo Koehler and Thomas Robinson and Tianhe Li and Tianjun Zhang and Tim Matthews and Timothy Chou and Tzook Shaked and Varun Vontimitta and Victoria Ajayi and Victoria Montanez and Vijai Mohan and Vinay Satish Kumar and Vishal Mangla and Vlad Ionescu and Vlad Poenaru and Vlad Tiberiu Mihailescu and Vladimir Ivanov and Wei Li and Wenchen Wang and Wenwen Jiang and Wes Bouaziz and Will Constable and Xiaocheng Tang and Xiaojian Wu and Xiaolan Wang and Xilun Wu and Xinbo Gao and Yaniv Kleinman and Yanjun Chen and Ye Hu and Ye Jia and Ye Qi and Yenda Li and Yilin Zhang and Ying Zhang and Yossi Adi and Youngjin Nam and Yu and Wang and Yu Zhao and Yuchen Hao and Yundi Qian and Yunlu Li and Yuzi He and Zach Rait and Zachary DeVito and Zef Rosnbrick and Zhaoduo Wen and Zhenyu Yang and Zhiwei Zhao and Zhiyu Ma},
      year={2024},
      eprint={2407.21783},
      archivePrefix={arXiv},
      primaryClass={cs.AI},
      url={https://arxiv.org/abs/2407.21783}, 
}

@inproceedings{boguraev2025causal,
    title = "Causal Interventions Reveal Shared Structure Across {E}nglish Filler{--}Gap Constructions",
    author = "Boguraev, Sasha  and
      Potts, Christopher  and
      Mahowald, Kyle",
    editor = "Christodoulopoulos, Christos  and
      Chakraborty, Tanmoy  and
      Rose, Carolyn  and
      Peng, Violet",
    booktitle = "Proceedings of the 2025 Conference on Empirical Methods in Natural Language Processing",
    month = nov,
    year = "2025",
    address = "Suzhou, China",
    publisher = "Association for Computational Linguistics",
    url = "https://aclanthology.org/2025.emnlp-main.1271/",
    doi = "10.18653/v1/2025.emnlp-main.1271",
    pages = "25021--25042",
    ISBN = "979-8-89176-332-6"
}

@inproceedings{gur2025mixing,
title={Mixing Mechanisms: How Language Models Retrieve Bound Entities In-Context},
author={Yoav Gur-Arieh and Mor Geva and Atticus Geiger},
booktitle={The Fourteenth International Conference on Learning Representations},
year={2026},
url={https://openreview.net/forum?id=UJ2UUjT2ko}
}

@inproceedings{wu-etal-2024-pyvene,
    title = "pyvene: A Library for Understanding and Improving {P}y{T}orch Models via Interventions",
    author = "Wu, Zhengxuan and Geiger, Atticus and Arora, Aryaman and Huang, Jing and Wang, Zheng and Goodman, Noah and Manning, Christopher and Potts, Christopher",
    editor = "Chang, Kai-Wei and Lee, Annie and Rajani, Nazneen",
    booktitle = "Proceedings of the 2024 Conference of the North American Chapter of the Association for Computational Linguistics: Human Language Technologies (Volume 3: System Demonstrations)",
    month = jun,
    year = "2024",
    address = "Mexico City, Mexico",
    publisher = "Association for Computational Linguistics",
    url = "https://aclanthology.org/2024.naacl-demo.16",
    pages = "158--165",
}

@inproceedings{kim2023entity,
    title = "Entity Tracking in Language Models",
    author = "Kim, Najoung  and
      Schuster, Sebastian",
    editor = "Rogers, Anna  and
      Boyd-Graber, Jordan  and
      Okazaki, Naoaki",
    booktitle = "Proceedings of the 61st Annual Meeting of the Association for Computational Linguistics (Volume 1: Long Papers)",
    month = jul,
    year = "2023",
    address = "Toronto, Canada",
    publisher = "Association for Computational Linguistics",
    url = "https://aclanthology.org/2023.acl-long.213/",
    doi = "10.18653/v1/2023.acl-long.213",
    pages = "3835--3855"
}

@inproceedings{chintam-etal-2023-identifying,
    title = "Identifying and Adapting Transformer-Components Responsible for Gender Bias in an {E}nglish Language Model",
    author = "Chintam, Abhijith  and
      Beloch, Rahel  and
      Zuidema, Willem  and
      Hanna, Michael  and
      van der Wal, Oskar",
    editor = "Belinkov, Yonatan  and
      Hao, Sophie  and
      Jumelet, Jaap  and
      Kim, Najoung  and
      McCarthy, Arya  and
      Mohebbi, Hosein",
    booktitle = "Proceedings of the 6th BlackboxNLP Workshop: Analyzing and Interpreting Neural Networks for NLP",
    month = dec,
    year = "2023",
    address = "Singapore",
    publisher = "Association for Computational Linguistics",
    url = "https://aclanthology.org/2023.blackboxnlp-1.29/",
    doi = "10.18653/v1/2023.blackboxnlp-1.29",
    pages = "379--394"
}

@misc{elhage2022toy,
      title={Toy Models of Superposition}, 
      author={Nelson Elhage and Tristan Hume and Catherine Olsson and Nicholas Schiefer and Tom Henighan and Shauna Kravec and Zac Hatfield-Dodds and Robert Lasenby and Dawn Drain and Carol Chen and Roger Grosse and Sam McCandlish and Jared Kaplan and Dario Amodei and Martin Wattenberg and Christopher Olah},
      year={2022},
      eprint={2209.10652},
      archivePrefix={arXiv},
      primaryClass={cs.LG},
      url={https://arxiv.org/abs/2209.10652}, 
}

@inproceedings{
hanna2024have,
title={Have Faith in Faithfulness: Going Beyond Circuit Overlap When Finding Model Mechanisms},
author={Michael Hanna and Sandro Pezzelle and Yonatan Belinkov},
booktitle={ICML 2024 Workshop on Mechanistic Interpretability},
year={2024},
url={https://openreview.net/forum?id=grXgesr5dT}
}

@inproceedings{
mueller2024missed,
title={Missed Causes and Ambiguous Effects: Counterfactuals Pose Challenges for Interpreting Neural Networks},
author={Aaron Mueller},
booktitle={ICML 2024 Workshop on Mechanistic Interpretability},
year={2024},
url={https://openreview.net/forum?id=pJs3ZiKBM5}
}

@inproceedings{saphra2024mechanistic,
    title = "Mechanistic?",
    author = "Saphra, Naomi  and
      Wiegreffe, Sarah",
    editor = "Belinkov, Yonatan  and
      Kim, Najoung  and
      Jumelet, Jaap  and
      Mohebbi, Hosein  and
      Mueller, Aaron  and
      Chen, Hanjie",
    booktitle = "Proceedings of the 7th BlackboxNLP Workshop: Analyzing and Interpreting Neural Networks for NLP",
    month = nov,
    year = "2024",
    address = "Miami, Florida, US",
    publisher = "Association for Computational Linguistics",
    url = "https://aclanthology.org/2024.blackboxnlp-1.30/",
    doi = "10.18653/v1/2024.blackboxnlp-1.30",
    pages = "480--498"
}

@inproceedings{brown2020language,
 author = {Brown, Tom and Mann, Benjamin and Ryder, Nick and Subbiah, Melanie and Kaplan, Jared D and Dhariwal, Prafulla and Neelakantan, Arvind and Shyam, Pranav and Sastry, Girish and Askell, Amanda and Agarwal, Sandhini and Herbert-Voss, Ariel and Krueger, Gretchen and Henighan, Tom and Child, Rewon and Ramesh, Aditya and Ziegler, Daniel and Wu, Jeffrey and Winter, Clemens and Hesse, Chris and Chen, Mark and Sigler, Eric and Litwin, Mateusz and Gray, Scott and Chess, Benjamin and Clark, Jack and Berner, Christopher and McCandlish, Sam and Radford, Alec and Sutskever, Ilya and Amodei, Dario},
 booktitle = {Advances in Neural Information Processing Systems},
 editor = {H. Larochelle and M. Ranzato and R. Hadsell and M.F. Balcan and H. Lin},
 pages = {1877--1901},
 publisher = {Curran Associates, Inc.},
 title = {Language Models are Few-Shot Learners},
 url = {https://proceedings.neurips.cc/paper_files/paper/2020/file/1457c0d6bfcb4967418bfb8ac142f64a-Paper.pdf},
 volume = {33},
 year = {2020}
}

@inproceedings{
prakash2024fine,
title={Fine-Tuning Enhances Existing Mechanisms: A Case Study on Entity Tracking},
author={Nikhil Prakash and Tamar Rott Shaham and Tal Haklay and Yonatan Belinkov and David Bau},
booktitle={The Twelfth International Conference on Learning Representations},
year={2024},
url={https://openreview.net/forum?id=8sKcAWOf2D}
}

@inproceedings{
feucht2025dual,
title={The Dual-Route Model of Induction},
author={Sheridan Feucht and Eric Todd and Byron C Wallace and David Bau},
booktitle={Second Conference on Language Modeling},
year={2025},
url={https://openreview.net/forum?id=bNTrKqqnG9}
}

@article{geiger2025causal,
  author  = {Atticus Geiger and Duligur Ibeling and Amir Zur and Maheep Chaudhary and Sonakshi Chauhan and Jing Huang and Aryaman Arora and Zhengxuan Wu and Noah Goodman and Christopher Potts and Thomas Icard},
  title   = {Causal Abstraction: A Theoretical Foundation for Mechanistic Interpretability},
  journal = {Journal of Machine Learning Research},
  year    = {2025},
  volume  = {26},
  number  = {83},
  pages   = {1--64},
  url     = {http://jmlr.org/papers/v26/23-0058.html}
}

@article{
sharkey2025open,
title={Open Problems in Mechanistic Interpretability},
author={Lee Sharkey and Bilal Chughtai and Joshua Batson and Jack Lindsey and Jeffrey Wu and Lucius Bushnaq and Nicholas Goldowsky-Dill and Stefan Heimersheim and Alejandro Ortega and Joseph Isaac Bloom and Stella Biderman and Adri{\`a} Garriga-Alonso and Arthur Conmy and Neel Nanda and Jessica Mary Rumbelow and Martin Wattenberg and Nandi Schoots and Joseph Miller and William Saunders and Eric J Michaud and Stephen Casper and Max Tegmark and David Bau and Eric Todd and Atticus Geiger and Mor Geva and Jesse Hoogland and Daniel Murfet and Thomas McGrath},
journal={Transactions on Machine Learning Research},
issn={2835-8856},
year={2025},
url={https://openreview.net/forum?id=91H76m9Z94}
}

@misc{DBLP:journals/corr/abs-2405-00208,
      title={A Primer on the Inner Workings of Transformer-based Language Models}, 
      author={Javier Ferrando and Gabriele Sarti and Arianna Bisazza and Marta R. Costa-jussà},
      year={2024},
      eprint={2405.00208},
      archivePrefix={arXiv},
      primaryClass={cs.CL},
      url={https://arxiv.org/abs/2405.00208}, 
}

@article{alishahi2019analyzing, title={Analyzing and interpreting neural networks for NLP: A report on the first BlackboxNLP workshop}, volume={25}, DOI={10.1017/S135132491900024X}, number={4}, journal={Natural Language Engineering}, author={Alishahi, Afra and Chrupała, Grzegorz and Linzen, Tal}, year={2019}, pages={543–557}}

@InProceedings{pmlr-v275-pislar25a,
  title = 	 {Combining Causal Models for More Accurate Abstractions of Neural Networks},
  author =       {P\^{i}slar, Theodora-Mara and Magliacane, Sara and Geiger, Atticus},
  booktitle = 	 {Proceedings of the Fourth Conference on Causal Learning and Reasoning},
  pages = 	 {114--138},
  year = 	 {2025},
  editor = 	 {Huang, Biwei and Drton, Mathias},
  volume = 	 {275},
  series = 	 {Proceedings of Machine Learning Research},
  month = 	 {07--09 May},
  publisher =    {PMLR},
  url = 	 {https://proceedings.mlr.press/v275/pislar25a.html}
}

@misc{gemmateam2024gemma2improvingopen,
      title={Gemma 2: Improving Open Language Models at a Practical Size}, 
      author={Gemma Team and Morgane Riviere and Shreya Pathak and Pier Giuseppe Sessa and Cassidy Hardin and Surya Bhupatiraju and Léonard Hussenot and Thomas Mesnard and Bobak Shahriari and Alexandre Ramé and Johan Ferret and Peter Liu and Pouya Tafti and Abe Friesen and Michelle Casbon and Sabela Ramos and Ravin Kumar and Charline Le Lan and Sammy Jerome and Anton Tsitsulin and Nino Vieillard and Piotr Stanczyk and Sertan Girgin and Nikola Momchev and Matt Hoffman and Shantanu Thakoor and Jean-Bastien Grill and Behnam Neyshabur and Olivier Bachem and Alanna Walton and Aliaksei Severyn and Alicia Parrish and Aliya Ahmad and Allen Hutchison and Alvin Abdagic and Amanda Carl and Amy Shen and Andy Brock and Andy Coenen and Anthony Laforge and Antonia Paterson and Ben Bastian and Bilal Piot and Bo Wu and Brandon Royal and Charlie Chen and Chintu Kumar and Chris Perry and Chris Welty and Christopher A. Choquette-Choo and Danila Sinopalnikov and David Weinberger and Dimple Vijaykumar and Dominika Rogozińska and Dustin Herbison and Elisa Bandy and Emma Wang and Eric Noland and Erica Moreira and Evan Senter and Evgenii Eltyshev and Francesco Visin and Gabriel Rasskin and Gary Wei and Glenn Cameron and Gus Martins and Hadi Hashemi and Hanna Klimczak-Plucińska and Harleen Batra and Harsh Dhand and Ivan Nardini and Jacinda Mein and Jack Zhou and James Svensson and Jeff Stanway and Jetha Chan and Jin Peng Zhou and Joana Carrasqueira and Joana Iljazi and Jocelyn Becker and Joe Fernandez and Joost van Amersfoort and Josh Gordon and Josh Lipschultz and Josh Newlan and Ju-yeong Ji and Kareem Mohamed and Kartikeya Badola and Kat Black and Katie Millican and Keelin McDonell and Kelvin Nguyen and Kiranbir Sodhia and Kish Greene and Lars Lowe Sjoesund and Lauren Usui and Laurent Sifre and Lena Heuermann and Leticia Lago and Lilly McNealus and Livio Baldini Soares and Logan Kilpatrick and Lucas Dixon and Luciano Martins and Machel Reid and Manvinder Singh and Mark Iverson and Martin Görner and Mat Velloso and Mateo Wirth and Matt Davidow and Matt Miller and Matthew Rahtz and Matthew Watson and Meg Risdal and Mehran Kazemi and Michael Moynihan and Ming Zhang and Minsuk Kahng and Minwoo Park and Mofi Rahman and Mohit Khatwani and Natalie Dao and Nenshad Bardoliwalla and Nesh Devanathan and Neta Dumai and Nilay Chauhan and Oscar Wahltinez and Pankil Botarda and Parker Barnes and Paul Barham and Paul Michel and Pengchong Jin and Petko Georgiev and Phil Culliton and Pradeep Kuppala and Ramona Comanescu and Ramona Merhej and Reena Jana and Reza Ardeshir Rokni and Rishabh Agarwal and Ryan Mullins and Samaneh Saadat and Sara Mc Carthy and Sarah Cogan and Sarah Perrin and Sébastien M. R. Arnold and Sebastian Krause and Shengyang Dai and Shruti Garg and Shruti Sheth and Sue Ronstrom and Susan Chan and Timothy Jordan and Ting Yu and Tom Eccles and Tom Hennigan and Tomas Kocisky and Tulsee Doshi and Vihan Jain and Vikas Yadav and Vilobh Meshram and Vishal Dharmadhikari and Warren Barkley and Wei Wei and Wenming Ye and Woohyun Han and Woosuk Kwon and Xiang Xu and Zhe Shen and Zhitao Gong and Zichuan Wei and Victor Cotruta and Phoebe Kirk and Anand Rao and Minh Giang and Ludovic Peran and Tris Warkentin and Eli Collins and Joelle Barral and Zoubin Ghahramani and Raia Hadsell and D. Sculley and Jeanine Banks and Anca Dragan and Slav Petrov and Oriol Vinyals and Jeff Dean and Demis Hassabis and Koray Kavukcuoglu and Clement Farabet and Elena Buchatskaya and Sebastian Borgeaud and Noah Fiedel and Armand Joulin and Kathleen Kenealy and Robert Dadashi and Alek Andreev},
      year={2024},
      eprint={2408.00118},
      archivePrefix={arXiv},
      primaryClass={cs.CL},
      url={https://arxiv.org/abs/2408.00118}, 
}

@misc{yang2025qwen3,
      title={Qwen3 Technical Report}, 
      author={An Yang and Anfeng Li and Baosong Yang and Beichen Zhang and Binyuan Hui and Bo Zheng and Bowen Yu and Chang Gao and Chengen Huang and Chenxu Lv and Chujie Zheng and Dayiheng Liu and Fan Zhou and Fei Huang and Feng Hu and Hao Ge and Haoran Wei and Huan Lin and Jialong Tang and Jian Yang and Jianhong Tu and Jianwei Zhang and Jianxin Yang and Jiaxi Yang and Jing Zhou and Jingren Zhou and Junyang Lin and Kai Dang and Keqin Bao and Kexin Yang and Le Yu and Lianghao Deng and Mei Li and Mingfeng Xue and Mingze Li and Pei Zhang and Peng Wang and Qin Zhu and Rui Men and Ruize Gao and Shixuan Liu and Shuang Luo and Tianhao Li and Tianyi Tang and Wenbiao Yin and Xingzhang Ren and Xinyu Wang and Xinyu Zhang and Xuancheng Ren and Yang Fan and Yang Su and Yichang Zhang and Yinger Zhang and Yu Wan and Yuqiong Liu and Zekun Wang and Zeyu Cui and Zhenru Zhang and Zhipeng Zhou and Zihan Qiu},
      year={2025},
      eprint={2505.09388},
      archivePrefix={arXiv},
      primaryClass={cs.CL},
      url={https://arxiv.org/abs/2505.09388}, 
}

@inproceedings{
olmo20242,
title={2 {OLM}o 2 Furious ({COLM}{\textquoteright}s Version)},
author={Evan Pete Walsh and Luca Soldaini and Dirk Groeneveld and Kyle Lo and Shane Arora and Akshita Bhagia and Yuling Gu and Shengyi Huang and Matt Jordan and Nathan Lambert and Dustin Schwenk and Oyvind Tafjord and Taira Anderson and David Atkinson and Faeze Brahman and Christopher Clark and Pradeep Dasigi and Nouha Dziri and Allyson Ettinger and Michal Guerquin and David Heineman and Hamish Ivison and Pang Wei Koh and Jiacheng Liu and Saumya Malik and William Merrill and Lester James Validad Miranda and Jacob Morrison and Tyler Murray and Crystal Nam and Jake Poznanski and Valentina Pyatkin and Aman Rangapur and Michael Schmitz and Sam Skjonsberg and David Wadden and Christopher Wilhelm and Michael Wilson and Luke Zettlemoyer and Ali Farhadi and Noah A. Smith and Hannaneh Hajishirzi},
booktitle={Second Conference on Language Modeling},
year={2025},
url={https://openreview.net/forum?id=2ezugTT9kU}
}

@inproceedings{ovalle-etal-2024-tokenization,
    title = "Tokenization Matters: Navigating Data-Scarce Tokenization for Gender Inclusive Language Technologies",
    author = "Ovalle, Anaelia  and
      Mehrabi, Ninareh  and
      Goyal, Palash  and
      Dhamala, Jwala  and
      Chang, Kai-Wei  and
      Zemel, Richard  and
      Galstyan, Aram  and
      Pinter, Yuval  and
      Gupta, Rahul",
    editor = "Duh, Kevin  and
      Gomez, Helena  and
      Bethard, Steven",
    booktitle = "Findings of the Association for Computational Linguistics: NAACL 2024",
    month = jun,
    year = "2024",
    address = "Mexico City, Mexico",
    publisher = "Association for Computational Linguistics",
    url = "https://aclanthology.org/2024.findings-naacl.113/",
    doi = "10.18653/v1/2024.findings-naacl.113",
    pages = "1739--1756"
}

@article{Zeldes_2017, title={The GUM corpus: creating multilayer resources in the classroom}, volume={51}, ISSN={1574-0218}, DOI={10.1007/s10579-016-9343-x}, abstractNote={This paper presents the methodology, design principles and detailed evaluation of a new freely available multilayer corpus, collected and edited via classroom annotation using collaborative software. After briefly discussing corpus design for open, extensible corpora, five classroom annotation projects are presented, covering structural markup in TEI XML, multiple part of speech tagging, constituent and dependency parsing, information structural and coreference annotation, and Rhetorical Structure Theory analysis. Layers are inspected for annotation quality and together they coalesce to form a richly annotated corpus that can be used to study the interactions between different levels of linguistic description. The evaluation gives an indication of the expected quality of a corpus created by students with relatively little training. A multifactorial example study on lexical NP coreference likelihood is also presented, which illustrates some applications of the corpus. The results of this project show that high quality, richly annotated resources can be created effectively as part of a linguistics curriculum, opening new possibilities not just for research, but also for corpora in linguistics pedagogy.}, number={3}, journal={Language Resources and Evaluation}, author={Zeldes, Amir}, year={2017}, month={sept}, pages={581–612} }

@misc{Weischedel-et-al-ontonotes,
title={OntoNotes Release 5.0}, url={https://catalog.ldc.upenn.edu/LDC2013T19}, DOI={10.35111/XMHB-2B84},
publisher={Linguistic Data Consortium}, author={Weischedel, Ralph and Palmer, Martha and Marcus, Mitchell and Hovy, Eduard and Pradhan, Sameer and Ramshaw, Lance and Xue, Nianwen and Taylor, Ann and Kaufman, Jeff and Franchini, Michelle and El-Bachouti, Mohammed and Belvin, Robert and Houston, Ann}, year={2013}, month=oct, pages={2806280 KB} }

@misc{chen2026circuitsleadromerethinking,
      title={All Circuits Lead to Rome: Rethinking Functional Anisotropy in Circuit and Sheaf Discovery for LLMs}, 
      author={Xi Chen and Mingyu Jin and Jingcheng Niu and Yutong Yin and Jinman Zhao and Bangwei Guo and Dimitris N. Metaxas and Zhaoran Wang and Yutao Yue and Gerald Penn},
      year={2026},
      eprint={2605.12671},
      archivePrefix={arXiv},
      primaryClass={cs.CL},
      url={https://arxiv.org/abs/2605.12671}, 
}

@inproceedings{gautam-2026-training,
    title = "Training in Step-by-Step Formal Reasoning Improves Pronominal Reasoning in Language Models",
    author = "Gautam, Vagrant",
    editor = "Demberg, Vera  and
      Inui, Kentaro  and
      Marquez, Llu{\'i}s",
    booktitle = "Proceedings of the 19th Conference of the {E}uropean Chapter of the {A}ssociation for {C}omputational {L}inguistics (Volume 2: Short Papers)",
    month = mar,
    year = "2026",
    address = "Rabat, Morocco",
    publisher = "Association for Computational Linguistics",
    url = "https://aclanthology.org/2026.eacl-short.7/",
    doi = "10.18653/v1/2026.eacl-short.7",
    pages = "121--135",
    ISBN = "979-8-89176-381-4"
}

@inproceedings{ovalle-et-al-2023-tango,
author = {Ovalle, Anaelia and Goyal, Palash and Dhamala, Jwala and Jaggers, Zachary and Chang, Kai-Wei and Galstyan, Aram and Zemel, Richard and Gupta, Rahul},
title = {“I’m fully who I am”: Towards Centering Transgender and Non-Binary Voices to Measure Biases in Open Language Generation},
year = {2023},
isbn = {9798400701924},
publisher = {Association for Computing Machinery},
address = {New York, NY, USA},
url = {https://doi.org/10.1145/3593013.3594078},
doi = {10.1145/3593013.3594078},
booktitle = {Proceedings of the 2023 ACM Conference on Fairness, Accountability, and Transparency},
pages = {1246–1266},
numpages = {21},
location = {Chicago, IL, USA},
series = {FAccT '23}
}

@inproceedings{hossain-etal-2023-misgendered,
    title = "{MISGENDERED}: Limits of Large Language Models in Understanding Pronouns",
    author = "Hossain, Tamanna  and
      Dev, Sunipa  and
      Singh, Sameer",
    editor = "Rogers, Anna  and
      Boyd-Graber, Jordan  and
      Okazaki, Naoaki",
    booktitle = "Proceedings of the 61st Annual Meeting of the Association for Computational Linguistics (Volume 1: Long Papers)",
    month = jul,
    year = "2023",
    address = "Toronto, Canada",
    publisher = "Association for Computational Linguistics",
    url = "https://aclanthology.org/2023.acl-long.293/",
    doi = "10.18653/v1/2023.acl-long.293",
    pages = "5352--5367"
}

@inproceedings{
subramonian2025agree,
title={Agree to Disagree? A Meta-Evaluation of {LLM} Misgendering},
author={Arjun Subramonian and Vagrant Gautam and Preethi Seshadri and Dietrich Klakow and Kai-Wei Chang and Yizhou Sun},
booktitle={Second Conference on Language Modeling},
year={2025},
url={https://openreview.net/forum?id=vgmiRvpCLA}
}

@misc{kotek2026protextbenchmarkdatasetmeasuring,
      title={ProText: A benchmark dataset for measuring (mis)gendering in long-form texts}, 
      author={Hadas Kotek and Margit Bowler and Patrick Sonnenberg and Yu'an Yang},
      year={2026},
      eprint={2603.27838},
      archivePrefix={arXiv},
      primaryClass={cs.CL},
      url={https://arxiv.org/abs/2603.27838}, 
}

@article{Guo_Yang_Zhang_Song_Wang_Zhu_Xu_Zhang_Ma_Bi_etal._2025, title={DeepSeek-R1 incentivizes reasoning in LLMs through reinforcement learning}, volume={645}, ISSN={1476-4687}, DOI={10.1038/s41586-025-09422-z}, abstractNote={General reasoning represents a long-standing and formidable challenge in artificial intelligence (AI). Recent breakthroughs, exemplified by large language models (LLMs)1,2 and chain-of-thought (CoT) prompting3, have achieved considerable success on foundational reasoning tasks. However, this success is heavily contingent on extensive human-annotated demonstrations and the capabilities of models are still insufficient for more complex problems. Here we show that the reasoning abilities of LLMs can be incentivized through pure reinforcement learning (RL), obviating the need for human-labelled reasoning trajectories. The proposed RL framework facilitates the emergent development of advanced reasoning patterns, such as self-reflection, verification and dynamic strategy adaptation. Consequently, the trained model achieves superior performance on verifiable tasks such as mathematics, coding competitions and STEM fields, surpassing its counterparts trained through conventional supervised learning on human demonstrations. Moreover, the emergent reasoning patterns exhibited by these large-scale models can be systematically used to guide and enhance the reasoning capabilities of smaller models.}, number={8081}, journal={Nature}, author={Guo, Daya and Yang, Dejian and Zhang, Haowei and Song, Junxiao and Wang, Peiyi and Zhu, Qihao and Xu, Runxin and Zhang, Ruoyu and Ma, Shirong and Bi, Xiao and Zhang, Xiaokang and Yu, Xingkai and Wu, Yu and Wu, Z. F. and Gou, Zhibin and Shao, Zhihong and Li, Zhuoshu and Gao, Ziyi and Liu, Aixin and Xue, Bing and Wang, Bingxuan and Wu, Bochao and Feng, Bei and Lu, Chengda and Zhao, Chenggang and Deng, Chengqi and Ruan, Chong and Dai, Damai and Chen, Deli and Ji, Dongjie and Li, Erhang and Lin, Fangyun and Dai, Fucong and Luo, Fuli and Hao, Guangbo and Chen, Guanting and Li, Guowei and Zhang, H. and Xu, Hanwei and Ding, Honghui and Gao, Huazuo and Qu, Hui and Li, Hui and Guo, Jianzhong and Li, Jiashi and Chen, Jingchang and Yuan, Jingyang and Tu, Jinhao and Qiu, Junjie and Li, Junlong and Cai, J. L. and Ni, Jiaqi and Liang, Jian and Chen, Jin and Dong, Kai and Hu, Kai and You, Kaichao and Gao, Kaige and Guan, Kang and Huang, Kexin and Yu, Kuai and Wang, Lean and Zhang, Lecong and Zhao, Liang and Wang, Litong and Zhang, Liyue and Xu, Lei and Xia, Leyi and Zhang, Mingchuan and Zhang, Minghua and Tang, Minghui and Zhou, Mingxu and Li, Meng and Wang, Miaojun and Li, Mingming and Tian, Ning and Huang, Panpan and Zhang, Peng and Wang, Qiancheng and Chen, Qinyu and Du, Qiushi and Ge, Ruiqi and Zhang, Ruisong and Pan, Ruizhe and Wang, Runji and Chen, R. J. and Jin, R. L. and Chen, Ruyi and Lu, Shanghao and Zhou, Shangyan and Chen, Shanhuang and Ye, Shengfeng and Wang, Shiyu and Yu, Shuiping and Zhou, Shunfeng and Pan, Shuting and Li, S. S. and Zhou, Shuang and Wu, Shaoqing and Yun, Tao and Pei, Tian and Sun, Tianyu and Wang, T. and Zeng, Wangding and Liu, Wen and Liang, Wenfeng and Gao, Wenjun and Yu, Wenqin and Zhang, Wentao and Xiao, W. L. and An, Wei and Liu, Xiaodong and Wang, Xiaohan and Chen, Xiaokang and Nie, Xiaotao and Cheng, Xin and Liu, Xin and Xie, Xin and Liu, Xingchao and Yang, Xinyu and Li, Xinyuan and Su, Xuecheng and Lin, Xuheng and Li, X. Q. and Jin, Xiangyue and Shen, Xiaojin and Chen, Xiaosha and Sun, Xiaowen and Wang, Xiaoxiang and Song, Xinnan and Zhou, Xinyi and Wang, Xianzu and Shan, Xinxia and Li, Y. K. and Wang, Y. Q. and Wei, Y. X. and Zhang, Yang and Xu, Yanhong and Li, Yao and Zhao, Yao and Sun, Yaofeng and Wang, Yaohui and Yu, Yi and Zhang, Yichao and Shi, Yifan and Xiong, Yiliang and He, Ying and Piao, Yishi and Wang, Yisong and Tan, Yixuan and Ma, Yiyang and Liu, Yiyuan and Guo, Yongqiang and Ou, Yuan and Wang, Yuduan and Gong, Yue and Zou, Yuheng and He, Yujia and Xiong, Yunfan and Luo, Yuxiang and You, Yuxiang and Liu, Yuxuan and Zhou, Yuyang and Zhu, Y. X. and Huang, Yanping and Li, Yaohui and Zheng, Yi and Zhu, Yuchen and Ma, Yunxian and Tang, Ying and Zha, Yukun and Yan, Yuting and Ren, Z. Z. and Ren, Zehui and Sha, Zhangli and Fu, Zhe and Xu, Zhean and Xie, Zhenda and Zhang, Zhengyan and Hao, Zhewen and Ma, Zhicheng and Yan, Zhigang and Wu, Zhiyu and Gu, Zihui and Zhu, Zijia and Liu, Zijun and Li, Zilin and Xie, Ziwei and Song, Ziyang and Pan, Zizheng and Huang, Zhen and Xu, Zhipeng and Zhang, Zhongyu and Zhang, Zhen}, year={2025}, month={sept}, pages={633–638} }

\clearpage

\appendix

\section{GPU hours} We ran approximately 150 GPU hours of experiments on NVIDIA A100 GPUs.

\section{Prompt design and Response Scoring}
\label{app:prompt-frame}
We frame the inputs as multiple-choice question to constrain the output space and thus make the counterfactual targets unambiguous.

\newtcolorbox{examplebox}{%
  enhanced,
  colback=gray!4,
  colframe=gray!4,
  borderline west={2pt}{0pt}{black!40},
  boxrule=0pt,
  arc=0pt,
  left=10pt, right=8pt, top=6pt, bottom=6pt,
  before skip=8pt, after skip=8pt
}

\begin{examplebox}
\textbf{Instruction:} Please select the correct pronoun from the options below.

\textbf{Input:} The \textbf{technician} thought that \textbf{he} had been having a bad day. The \textbf{technician} said that \textbf{\_\_\_} had completed the repair.

\textbf{Question:} What pronoun should be used to fill the blank? 

\textbf{Options:} he, she, they (randomized)

\textbf{Expected (Base):} he \quad 

\textbf{Expected ( \ModelG\ Intervention):} she

\end{examplebox}

The \textit{response token} is the first token the model generates after \texttt{Response:}, i.e., the token at the final position of the prompt prefix. Log-likelihoods are computed over the three candidate nominative pronouns \textit{he}, \textit{she}, and \textit{they}, each tokenized as a single token, and the candidate with the highest log-likelihood is taken as the prediction. All DAS interventions and mixture model distribution measurements are evaluated at this same position.

\section{Causal Models}
\label{app:causal_models_explained}
$O$ ranges over the 60 occupations in RUFF (e.g., \textit{technician}, \textit{nurse}); $P$ and $P'$ take values in \{\textit{he}, \textit{she}, \textit{they}\}; $S$ is the stereotypical pronoun associated with $O$ under the model's priors (all occupational biases, prompted with the task sentences of RUFF, are also listed in Table \ref{tab:all_models_occupation_bias}), also in \{\textit{he}, \textit{she}, \textit{they}\}; $E$ is a latent entity representation (not directly observed); and $Y \in \{\textit{he}, \textit{she}, \textit{they}\}$ is the predicted output pronoun.

\section{DAS Training Details}
\label{app:das-training}

Each DAS model learns an orthogonal rotation matrix $R \in \mathbb{R}^{d \times d}$ and soft boundaries over the residual stream at the identified best layer.
Each model is trained on all datasets (\texttt{GR, GS, RS}) separately, producing nine trained rotation matrices per model after full training. We jointly optimize $R$ and the boundary params using Adam with learning rates $10^{-3}$ and $10^{-2}$ respectively, with linear warmup over the first 10\% of steps, followed by linear decay. The boundary temperature, $\beta$ is annealed from 50 to 0.1 to encourage the boundaries to converge to a discrete subspace \cite{wu2023interpretability}. Models are trained for three epochs with an effective batch of 64 (\texttt{batch\_size=4}, \texttt{gradient\_accumulation\_steps=16}). We use an 80/20 train/eval split. The training objective is cross-entropy between the logits after intervention and the counterfactual label for the high-level mechanisms after intervention. We evaluate at the response token (see Appendix \ref{app:prompt-frame}).
We implement the experiments using the pyvene library \cite{wu-etal-2024-pyvene}.

\begin{table*}[t]
\section{Base Performances across Models}
\label{app:base-performance}
\centering
\small
\setlength{\tabcolsep}{2.5pt}
\begin{tabularx}{\linewidth}{Xcccccccccc}
\toprule
 & \multicolumn{2}{c}{\textbf{Overall}} & \multicolumn{3}{c}{\textbf{By Pronoun}} & \multicolumn{3}{c}{\textbf{By Case}} \\
\cmidrule(lr){2-3} \cmidrule(lr){4-6} \cmidrule(lr){7-9}
\textbf{Model} & \textbf{0-dist} & \textbf{1-dist} & \textbf{he} & \textbf{she} & \textbf{they} & \textbf{NOM} & \textbf{ACC} & \textbf{POSS} \\
\midrule
Llama-3.1-8B   & 83.2 & 51.2 & 88.6 / 33.3 & 61.1 / 24.4 & 100.0 / 97.0 & 66.7 / 47.9 & 95.9 / 56.3 & 88.7 / 30.0 \\
Llama-3.1-70B  & 90.4 & 48.8 & 91.0 / 51.2 & 90.4 / 33.3 &  89.8 / 62.2 & 95.0 / 67.9 & 84.1 / 31.3 & 92.0 / 30.0 \\
OLMo-2-1B      & 55.4 & 42.8 & 78.4 / 62.5 & 72.5 / 61.9 &  15.1 /  3.1 & 45.6 / 40.8 & 55.3 / 42.5 & 67.3 / 70.0 \\
OLMo-2-7B      & 69.6 & 46.0 & 98.2 / 98.8 & 79.6 / 33.3 &  30.7 /  4.9 & 70.0 / 49.6 & 72.4 / 42.1 & 66.0 / 50.0 \\
OLMo-2-13B     & 71.4 & 63.6 & 83.2 / 82.7 & 81.4 / 79.8 &  49.4 / 27.4 & 83.9 / 76.3 & 62.9 / 53.3 & 66.0 / 35.0 \\
OLMo-2-32B     & 85.4 & 51.6 & 83.2 / 41.7 & 92.2 / 45.8 &  80.7 / 67.7 & 76.7 / 64.2 & 91.8 / 38.8 & 88.7 / 55.0 \\
Gemma-2-9B     & 93.6 & 80.4 & 95.8 / 88.7 & 90.4 / 71.4 &  94.6 / 81.1 & 97.2 / 95.8 & 85.9 / 69.2 & 98.0 / 30.0 \\
Gemma-2-27B    & 94.2 & 74.2 & 98.2 / 78.6 & 99.4 / 86.3 &  84.9 / 57.3 & 99.4 / 100.0 & 83.5 / 48.8 & 100.0 / 70.0 \\
Qwen2.5-7B     & 52.4 & 43.2 & 90.4 / 90.5 & 53.3 / 34.5 &  13.3 /  3.7 & 49.4 / 40.4 & 62.4 / 48.8 & 44.7 / 10.0 \\
Qwen2.5-72B    & 93.2 & 82.0 & 96.4 / 78.0 & 97.6 / 84.5 &  85.5 / 83.5 & 97.8 / 100.0 & 92.9 / 69.6 & 88.0 / 15.0 \\
\bottomrule
\end{tabularx}
\caption{Base pronoun resolution accuracy (\%) per model. Pronoun and case columns show 0-dist / 1-dist.}
\label{tab:base_performance}
\end{table*}

\begin{figure*}[t]
\section{Layer Search Results}
\label{app:layer-search-all}
    \centering

    \begin{subfigure}{0.32\textwidth}
        \centering
        \includegraphics[width=\linewidth]{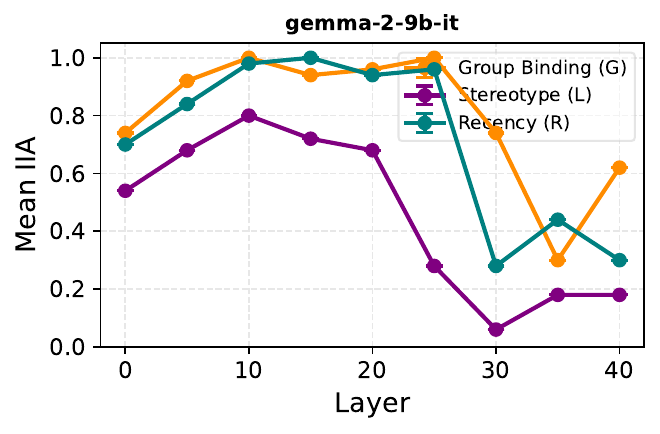}
        \caption{}
        \label{fig:gemma}
    \end{subfigure}
    \hfill
    \begin{subfigure}{0.32\textwidth}
        \centering
        \includegraphics[width=\linewidth]{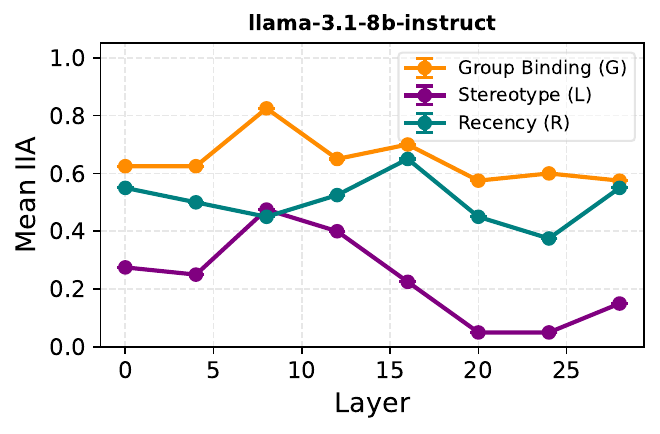}
        \caption{}
        \label{fig:llama}
    \end{subfigure}
    \hfill
    \begin{subfigure}{0.32\textwidth}
        \centering
        \includegraphics[width=\linewidth]{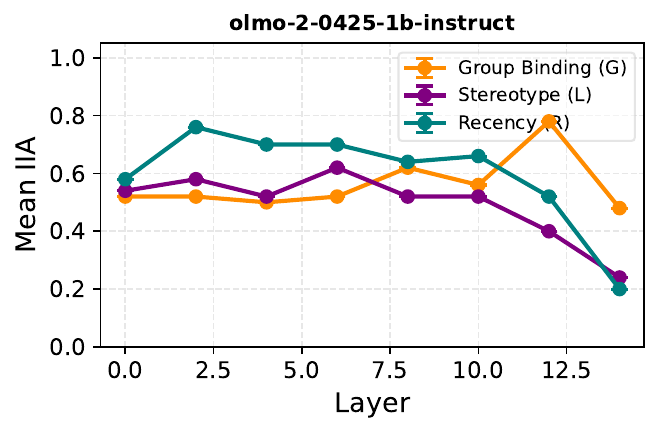}
        \caption{}
        \label{fig:olmo1b}
    \end{subfigure}

    \vspace{0.5em}

    \begin{subfigure}{0.32\textwidth}
        \centering
        \includegraphics[width=\linewidth]{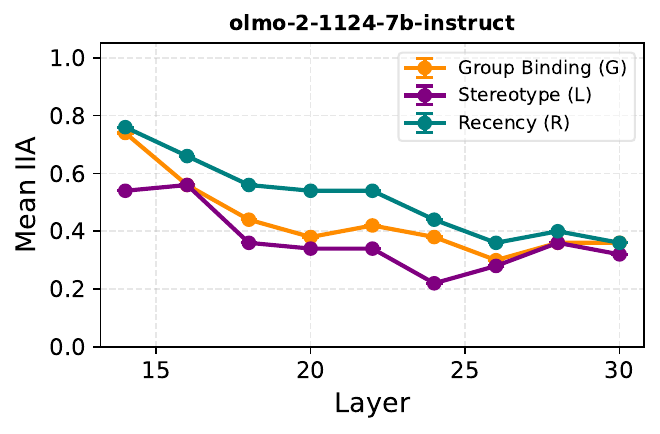}
        \caption{}
        \label{fig:olmo7b}
    \end{subfigure}
    \hfill
    \begin{subfigure}{0.32\textwidth}
        \centering
        \includegraphics[width=\linewidth]{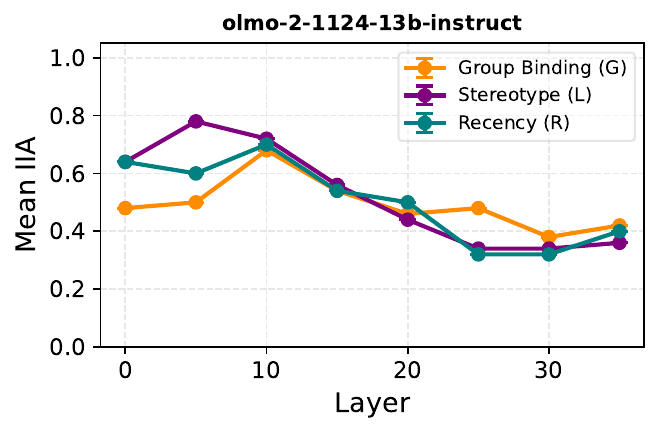}
        \caption{}
        \label{fig:olmo13b}
    \end{subfigure}
    \hfill
    \begin{subfigure}{0.32\textwidth}
        \centering
        \includegraphics[width=\linewidth]{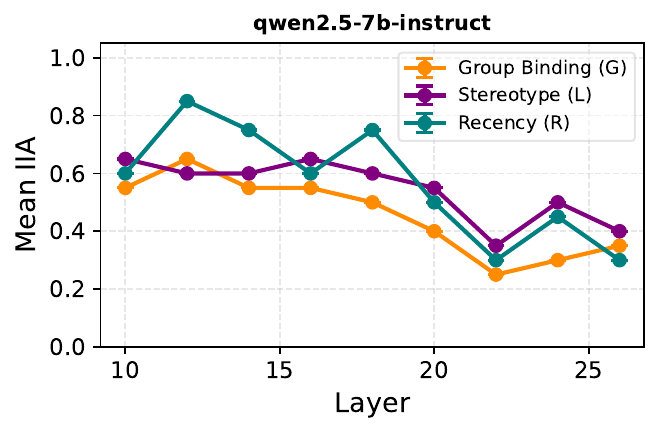}
        \caption{}
        \label{fig:qwen}
    \end{subfigure}

    \caption{Layer search comparison.}
    \label{fig:layer-search-all}
\end{figure*}

\begin{figure*}[p]
\section{Attention Head Experiments}
\label{app:attention-heads}

    \centering

    \begin{minipage}{0.48\textwidth}
        \centering
        \includegraphics[width=\linewidth]{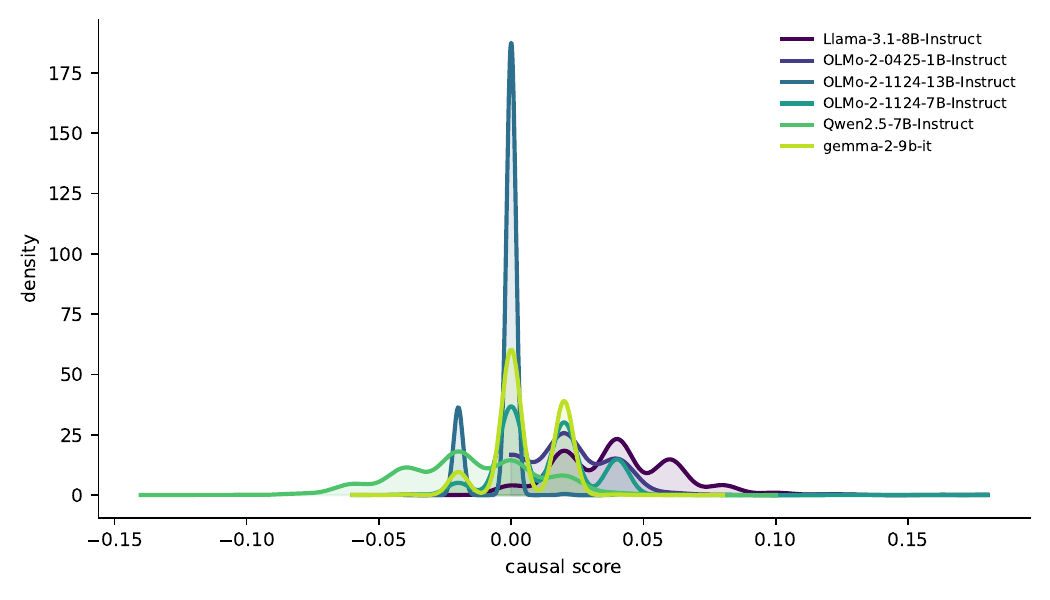}
        \caption{Causal scores for concept-level interventions.}
        \label{fig:causal-scores-concept}
    \end{minipage}
    \hfill
    \begin{minipage}{0.48\textwidth}
        \centering
        \includegraphics[width=\linewidth]{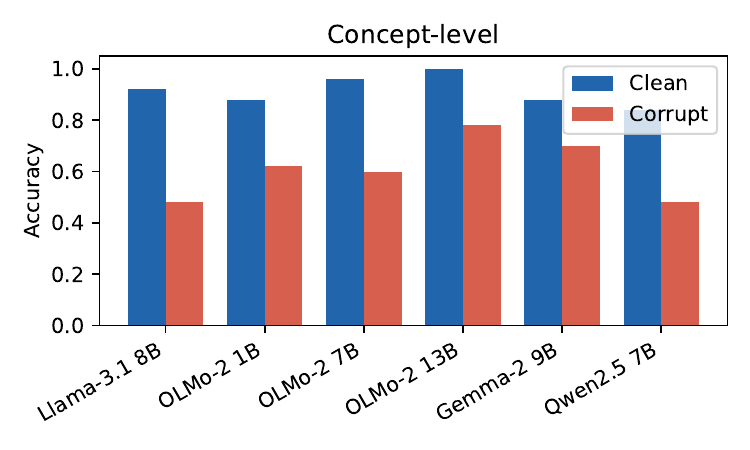}
        \caption{Activation patching results at the concept level.}
        \label{fig:patching-concept}
    \end{minipage}

\end{figure*}
\begin{figure*}[p]
    \centering

    \begin{minipage}{0.48\textwidth}
        \centering
        \includegraphics[width=\linewidth]{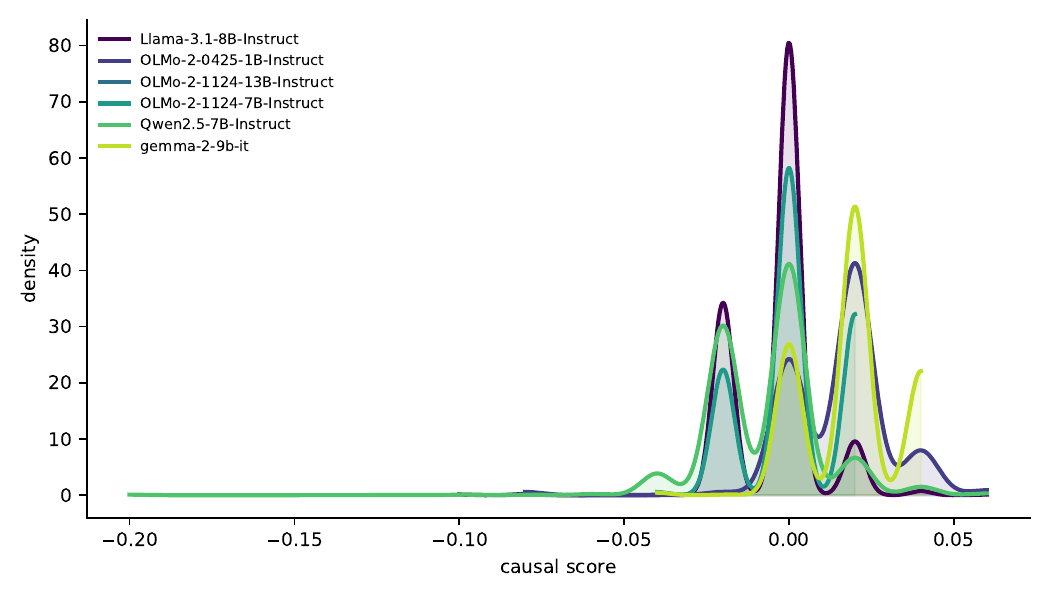}
        \caption{Causal scores for token-level interventions.}
        \label{fig:causal-scores-token}
    \end{minipage}
    \hfill
    \begin{minipage}{0.48\textwidth}
        \centering
        \includegraphics[width=\linewidth]{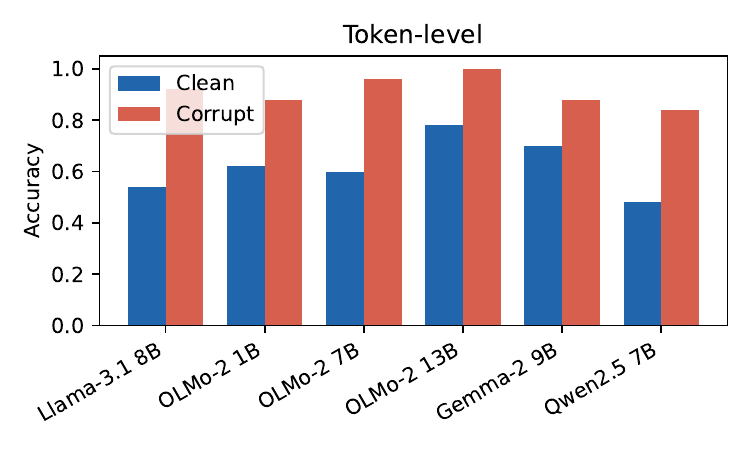}
        \caption{Activation patching results at the token level.}
        \label{fig:patching-token}
    \end{minipage}

\end{figure*}

\begin{table*}[t]
\section{Context-free Bias}
\centering
\small
\setlength{\tabcolsep}{2pt}
\begin{tabular}{lcccccccccc}
\hline
\textbf{Occ.} &
\textbf{Llama-8B} &
\textbf{-70B} &
\textbf{Olmo-1B} &
\textbf{-7B} &
\textbf{-13B} &
\textbf{-32B} &
\textbf{Qwen-7B} &
\textbf{-70B} &
\textbf{Gemma-9B} &
\textbf{-27B} \\
\hline

accountant   & he & they & he & she & he   & he   & he & he & he & he \\
administrator& he & they & he & she & he   & he   & he & he & he & he \\
advisor      & he & they & he & she & he   & he   & he & he & he & he \\
appraiser    & he & they & he & she & he   & he   & he & he & he & he \\
architect    & he & they & he & she & he   & he   & he & he & he & he \\
auditor      & he & they & he & she & he   & he   & he & he & he & he \\
baker        & he & they & he & she & he   & he   & he & he & he & he \\
bartender    & he & they & he & she & he   & he   & he & he & he & he \\
broker       & they & they & he & she & he  & he   & he & he & he & he \\
carpenter    & he & they & he & he  & he   & he   & he & he & he & he \\
cashier      & she & she  & he & she & they & he   & he & she & he & he \\
chef         & he & they & he & she & he   & he   & he & he & he & he \\
chemist      & he & they & he & she & he   & he   & he & he & he & he \\
clerk        & she & they & he & she & he   & he   & he & he & he & he \\
counselor    & she & they & he & she & they & he   & he & he & he & she \\
dietitian    & she & she  & he & she & she  & she  & he & she & she & she \\
dispatcher   & she & they & he & she & he   & he   & he & he & he & he \\
doctor       & he  & they & he & she & he   & he   & he & he & he & he \\
educator     & she & they & he & she & they & he   & he & he & he & she \\
electrician  & he  & they & he & he  & he   & he   & he & he & he & he \\
engineer     & he  & they & he & she & he   & he   & he & he & he & he \\
examiner     & he  & they & he & she & they & he   & he & he & he & he \\
firefighter  & he  & they & he & she & he   & he   & he & he & he & he \\
hairdresser  & she & she  & he & she & they & he   & he & he & she & she \\
hygienist    & she & she  & he & she & she  & she  & he & she & she & she \\
inspector    & he  & they & he & she & he   & he   & he & he & he & he \\
instructor   & she & they & he & she & he   & he   & he & he & he & he \\
investigator & he  & they & he & she & he   & he   & he & he & he & he \\
janitor      & he  & they & he & she & he   & he   & he & he & he & he \\
lawyer       & he  & they & he & she & he   & he   & he & he & he & he \\
librarian    & she & she  & he & she & they & he   & he & she & she & she \\
machinist    & he  & they & he & she & he   & he   & he & he & he & he \\
manager      & they & they & he & she & he   & he   & he & he & he & he \\
mechanic     & he  & they & he & he  & he   & he   & he & he & he & he \\
nurse        & she & she  & he & she & she  & she  & she & she & she & she \\
nutritionist & she & she  & he & she & they & he   & he & she & she & she \\
officer      & he  & they & he & she & he   & he   & he & he & he & he \\
painter      & he  & they & he & she & he   & he   & he & he & he & he \\
paralegal    & she & she  & he & she & they & he   & he & he & she & she \\
paramedic    & they & they & he & he  & he   & he   & he & he & he & he \\
pathologist  & he  & they & he & she & he   & he   & he & he & he & he \\
pharmacist   & he  & they & he & she & he   & he   & he & he & he & he \\
physician    & he  & they & he & she & he   & he   & he & he & he & he \\
planner      & she & they & he & she & they & he   & he & he & he & he \\
plumber      & he  & they & he & he  & he   & he   & he & he & he & he \\
practitioner & they & they & he & she & they & he   & he & he & he & he \\
programmer   & he  & they & he & she & he   & he   & he & he & he & he \\
psychologist & she & they & he & she & he   & he   & he & he & he & he \\
receptionist & she & she  & he & she & she  & she  & he & she & she & she \\
salesperson  & they & they & he & she & they & he   & he & he & he & he \\
scientist    & he  & they & he & she & he   & he   & he & he & he & he \\
secretary    & she & she  & he & she & he   & he   & he & she & she & she \\
specialist   & he  & they & he & she & he   & he   & he & he & he & he \\
supervisor   & he  & they & he & she & he   & he   & he & he & he & he \\
surgeon      & he  & they & he & she & he   & he   & he & he & he & he \\
teacher      & she & she  & he & she & they & he   & he & she & she & he \\
technician   & he  & they & he & she & he   & he   & he & he & he & he \\
therapist    & she & she  & he & she & they & he   & he & she & she & she \\
veterinarian & she & she  & he & she & he   & he   & he & he & he & he \\
worker       & they & they & he & she & he   & he   & he & he & he & he \\

\hline
\end{tabular}
\caption{Average over three runs: Most likely pronoun predictions across all models for occupations in the RUFF dataset. These are used to calculate $IIA_\texttt{stereotype}$ and to train \ModelS~for each language model.}
\label{tab:all_models_occupation_bias}
\end{table*}

\end{document}